\documentclass[journal]{IEEEtran}

\usepackage{algpseudocode}
\usepackage{algorithm}
\usepackage{amsmath}
\usepackage{amssymb}
\usepackage{booktabs}
\usepackage[justification=centering]{caption}
\usepackage{cite}
\usepackage{colortbl}
\usepackage{csquotes}
\usepackage{graphicx}
\usepackage{hyperref}
\usepackage[utf8]{inputenc}
\usepackage{lineno}
\usepackage{multirow}
\usepackage{orcidlink}
\usepackage{pgfplots}
\pgfplotsset{compat=1.8}
\usepackage{tikz}
\usetikzlibrary{arrows,arrows.meta,chains,decorations.markings,matrix,patterns,positioning,shapes.geometric,scopes,shadows,shadows.blur,trees}
\usepackage{xcolor}

\def\BibTeX{{\rm B\kern-.05em{\sc i\kern-.025em b}\kern-.08em
    T\kern-.1667em\lower.7ex\hbox{E}\kern-.125emX}}

\begin{document}

\title{Inadequacies of Large Language Model Benchmarks in the Era of Generative Artificial Intelligence}

\author{Timothy R. McIntosh\orcidlink{0000-0003-0836-4266}*, Teo Susnjak\orcidlink{0000-0001-9416-1435}, Nalin Arachchilage\orcidlink{0000-0002-0059-0376}, Tong Liu\orcidlink{0000-0003-3047-1148}, Dan Xu\orcidlink{0009-0004-3930-7381}, Paul Watters\orcidlink{0000-0002-1399-7175}, \IEEEmembership{Senior Member, IEEE}, and Malka N. Halgamuge\orcidlink{0000-0001-9994-3778}, \IEEEmembership{Senior Member, IEEE}
	\thanks{Manuscript received \protect\today. Corresponding Author: Timothy R. McIntosh (e-mail: timothy.mcintosh@rmit.edu.au).}
}

\maketitle

\begin{abstract}
	The rapid rise in popularity of Large Language Models (LLMs) with emerging capabilities has spurred public curiosity to evaluate and compare different LLMs, leading many researchers to propose their own LLM benchmarks. Noticing preliminary inadequacies in those benchmarks, we embarked on a study to critically assess 23 state-of-the-art LLM benchmarks, using our novel unified evaluation framework through the lenses of people, process, and technology, under the pillars of benchmark functionality and integrity. Our research uncovered significant limitations, including biases, difficulties in measuring genuine reasoning, adaptability, implementation inconsistencies, prompt engineering complexity, evaluator diversity, and the overlooking of cultural and ideological norms in one comprehensive assessment. Our discussions emphasized the urgent need for standardized methodologies, regulatory certainties, and ethical guidelines in light of Artificial Intelligence (AI) advancements, including advocating for an evolution from static benchmarks to dynamic behavioral profiling to accurately capture LLMs' complex behaviors and potential risks. Our study highlighted the necessity for a paradigm shift in LLM evaluation methodologies, underlining the importance of collaborative efforts for the development of universally accepted benchmarks and the enhancement of AI systems' integration into society.
\end{abstract}



\section{Introduction}
\label{sec:Introduction}
\IEEEPARstart{L}{arge} Language Models (LLMs), a sophisticated branch of generative Artificial Intelligence (AI), have evolved from narrow, task-specific systems to versatile models capable of few-shot learning and handling diverse tasks \cite{min2023recent,mcintosh2023google}. Evaluating LLMs is crucial for understanding their capabilities and limitations. Automatic evaluation methods, which employ standard metrics, offer computational efficiency, while human evaluations provide nuanced insights into the quality and accuracy of LLM responses \cite{chang2023survey,mcintosh2024reasoning}. Recent advancements like GPT-4 and Gemini, with their multimodal capabilities, and Mistral 8x7B's integration of Mixture of Experts (MoE), have enhanced LLMs' ability to process specialized domain knowledge and reason across diverse tasks \cite{chang2023survey,min2023recent,mcintosh2023google}. The proliferation of over 700,000 LLMs on platforms like HuggingFace, alongside many publicly available commercial LLMs, has intensified competition among developers, driving the need for benchmarks to uniformly evaluate and compare LLM performance across dimensions such as accuracy, robustness, and reasoning, as these directly influence an LLM’s reliability in real-world applications \cite{chang2023survey}. For example, an LLM that excels in accuracy but lacks robustness might fail when confronted with unexpected inputs or novel scenarios. Benchmarks serve as standardized sets of tasks or datasets that assess key aspects such as accuracy, efficiency, and ethical considerations like bias or fairness, guiding both development and deployment decisions \cite{chang2023survey}. Widely used benchmarks, such as GLUE, SuperGLUE, and MMLU, are expected to enable consistent LLM evaluations, allowing researchers to fine-tune LLMs for specific tasks or domains, and facilitating performance comparisons across different models in real-world scenarios.

Unlike the automobile and aviation industries, where clear regulations and well-defined public consensus guide benchmarking practices \cite{liu2015benchmarking}, the advanced AI field lacks such universally accepted standards, leading many researchers to devise their own benchmarks. Benchmarks of LLMs, such as BIG-bench \cite{srivastava2022beyond} by Google DeepMind and PromptBench \cite{zhu2024promptbench} by Microsoft Research, have encompassed diverse methods, each with unique approaches and criteria, focusing mainly on exam-style or task-based assessments. Such methods, evaluating models' abilities to perform specific functions or solve problems, typically emphasized tasks with predefined answers or scenarios with limited variability \cite{chen2021evaluating,zhu2024promptbench}. Models were evaluated using metrics like accuracy, perplexity, and F1-score on fixed datasets (\textit{e.g.}, \cite{liu2023m3ke,singhal2023large}), or using human evaluators to measure \enquote{human-level} performance (\textit{e.g.}, \cite{yu2023kola,zhong2024agieval}). The common approach of LLM benchmarking has often assumed a standardized \enquote{correct} answer or a marking rubric for each question, which LLMs are expected to reproduce, and the same benchmark sets are repeatedly used to compare and rank LLMs, ignoring their broader implications and real-world applicability. While providing insights, we believe this approach often fails to capture the subtleties and complexities of real-world use, as previous studies mainly focused on functionality within the technological context, neglecting processual and human aspects, especially the diversity of human values and cultures \cite{mcintosh2023culturally}. For instance, an LLM that performs well on standardized tasks may still struggle in contexts requiring cultural sensitivity, such as understanding sophisticated social interactions or ethical concerns in decision-making, which can vary significantly across different societies and user groups. Additionally, LLM benchmarks have often lacked a thorough examination, especially in evaluating LLMs' full capabilities, limitations, and safety issues \cite{guha2024legalbench}, highlighting the need for a more comprehensive and sophisticated approach. Evaluating LLMs and generative AI requires a method that assesses not only standardized task performance but also real-world applicability and safety, as a lack of such holistic evaluations can lead to LLMs that perform well in controlled environments but fail in critical real-world applications, posing risks such as perpetuating bias, making unsafe decisions, or being vulnerable to manipulation \cite{chang2023survey,mcintosh2023google}.

This study was motivated by our concern over the rapid proliferation of LLM benchmark studies and the competition among them to establish superiority, often at the expense of well-rounded functionality and integrity assurance. When evaluating LLM benchmarks, two core aspects are critical for a reliable and unbiased evaluation of LLMs: \textit{functionality}, which refers to how well a benchmark measures the specific capabilities of an LLM in alignment with real-world applications, and \textit{integrity}, which ensures that the benchmark resists manipulation or gaming by models that exploit its criteria to produce misleading results. An \textit{inadequacy}, in this context, refers to any deficiency or shortcoming in a benchmark's ability to fully capture an LLM's functionality or maintain its integrity during evaluation. Our preliminary observations indicated systemic inadequacies in most current LLM benchmarks, where superficial metrics often replaced comprehensive evaluations of LLM functionality and the assurance of benchmark integrity. As a result, many benchmarks failed to measure the complex, evolving capabilities of LLMs in real-world settings or address the risks of LLMs gaming and overfitting them. We propose a re-evaluation of current LLM benchmarking criteria, hypothesizing that a comprehensive evaluation framework must integrate both functionality and integrity, to provide a balanced, holistic, in-depth assessment of LLMs. This study aims to systematically analyze the inadequacies of current exam-style benchmarking in multimodal LLMs through the following research questions: (1) How can we identify, categorize, and explain the common inadequacies of state-of-the-art LLM benchmarks? (2) Are these inadequacies evident in the popular benchmarks we have identified? (3) What should a comprehensive evaluation of LLM benchmarks include, considering both functionality and integrity for a complete understanding of LLM risks and societal impacts? Through this survey study, we propose a unified evaluation framework for LLM benchmarks, aligned with the domains of people, process, and technology, designed to facilitate a thorough examination of benchmarks. Our framework aims to enhance both functionality and integrity in benchmark design, guiding the development of more effective and secure LLM evaluation methods that reflect real-world applicability and ensure robust model development and deployment.

The major contributions of this study are as follows:
\begin{enumerate}
	\item We proposed a unified evaluation framework for LLM benchmarks, based on the domains of people, process, and technology, as a foundation to comprehensively and holistically assess both the functionality and integrity of LLMs in real-world applications.
	\item Using this framework, we conducted a systematic critique of 23 state-of-the-art LLM benchmarks, revealing key inadequacies and offering targeted methodological improvements to enhance the accuracy and fairness of LLM assessments.
	\item Building on the insights from our framework, we introduced an advanced evaluation method that extended traditional benchmarking with behavioral profiling and regular audits, creating a dynamic and ongoing assessment process that addressed evolving issues of real-world applicability, inclusivity, and security.
\end{enumerate}

The rest of this paper is organized as follows: Section \ref{sec:background} offers a comprehensive analysis of prevalent benchmarks. Section \ref{sec:unified-evaluation-framework} introduces the proposed unified evaluation framework for assessing LLMs. Section \ref{sec:OverviewLLMBenchmarks} provides a detailed overview of 23 selected LLM benchmarks, summarizing their characteristics, focus areas, and evaluation methodologies. Section \ref{sec:technological-aspects} examines the technological challenges in current LLM benchmarking practices. Section \ref{sec:processual-elements} discusses the processual elements affecting LLM evaluations. Section \ref{sec:human-dynamics} explores the human dynamics influencing LLM benchmarking. Section \ref{sec:discussions} provides a discussion on the overarching themes and implications of our findings. Finally, Section \ref{sec:conclusion} concludes the paper, summarizing the key findings and contributions.

\section{Background and Related Work}
\label{sec:background}

Benchmarking is a critical process in computer science, serving as a standardized method to evaluate and compare the performance of hardware and software systems \cite{papadopoulos2019methodological}. This section reviews traditional benchmarking practices in computer science and contrasts them with the emerging challenges of benchmarking in generative AI and LLMs.

\subsection{Traditional Benchmarking in Computer Science}
\label{subsec:traditional-benchmarking}

In computer science, benchmarking involves running a set of standardized tests on hardware or software systems to measure their performance under controlled conditions \cite{papadopoulos2019methodological}. Such benchmarks are designed to be repeatable and objective, allowing for fair comparisons between different systems or configurations. For instance, the SPEC CPU benchmark suite evaluates a processor's ability to handle compute-intensive tasks by measuring execution times of standardized workloads \cite{brunst2022first}. Standardization is essential in benchmarking to ensure consistency across evaluations, making the results reliable and widely accepted within the community \cite{li2018benchmarking}. Benchmarking methodologies typically focus on specific performance metrics such as processing speed, memory bandwidth, or energy efficiency \cite{hort2021survey,papadopoulos2019methodological}. Tools like Cinebench and 3DMark assess computational performance by performing intensive CPU or GPU tasks, which is critical for high-performance computing applications, and they help identify system bottlenecks and guide optimization efforts, contributing to advancements in hardware and software design \cite{li2018benchmarking,hort2021survey,papadopoulos2019methodological}. The development of benchmarks requires careful consideration of workload representativeness to ensure that the tests reflect real-world usage scenarios \cite{aslanpour2020performance}.

However, traditional benchmarking faces challenges like benchmark manipulation, where systems are engineered to perform exceptionally well on specific benchmarks without delivering proportional real-world performance \cite{xia2020interactive}. For example, hardware manufacturers might optimize compilers or system settings to inflate benchmark scores, leading to misleading conclusions \cite{romano2021empirical}. This has prompted calls for more robust and comprehensive benchmarking practices that can resist gaming and provide a true measure of system capabilities \cite{davis2023benchmarks}. Additionally, the reliability of benchmarking results in computer science and AI depends critically on their reproducibility, generalizability across different contexts, consistency over time, and objectivity in measurement, but at the same time, it raises the question of who holds the authority to set benchmark standards \cite{davis2023benchmarks,li2018benchmarking}. Consider that the same anti-malware products can score differently in tests like VB100, AV-Comparatives, and AV-Test, even when tests are administered around the same time, indicating variations in benchmarking criteria and evaluation methodologies \cite{zhu2020measuring}. Moreover, within the context of cybersecurity, our previous survey on ransomware revealed concerns about the self-benchmarking practices in many anti-ransomware studies, which often used diverse ransomware samples and methodologies and based their claims of superiority merely on higher detection rates \cite{mcintosh2021ransomware}. This situation underscores the need for rigorous, comprehensive, and scientifically sound benchmarking practices that can provide reliable, meaningful and universally accepted comparisons.

\subsection{Benchmarking in Generative AI and LLMs}
\label{subsec:ai-benchmarking}

Benchmarking in generative AI and LLMs introduces unique challenges that differ fundamentally from traditional computer science benchmarking. LLMs like Llama and GPT-4 are capable of understanding and generating human-like text, performing tasks ranging from translation to creative writing \cite{min2023recent,mcintosh2024reasoning}. Evaluating such models requires benchmarks that can assess not just quantitative performance, but also qualitative aspects like coherence, relevance, and ethical considerations \cite{chang2023survey}. Unlike traditional benchmarks with clear numerical metrics, LLM benchmarking often involves subjective judgments and complex evaluation criteria \cite{van2024field,mcintosh2023culturally}. LLM benchmarks commonly involve tasks such as language understanding, reasoning, and dialogue generation, using datasets like GLUE to provide standardized evaluation platforms \cite{wang2019glue}. However, the open-ended nature of language means that there can be multiple valid responses to a given prompt, complicating the assessment of correctness \cite{min2023recent}. Furthermore, LLMs operate as black boxes with opaque internal mechanisms, making it difficult to interpret their decision-making processes and identify potential biases or ethical issues \cite{tan2024sparsity,mcintosh2024reasoning}. This opacity poses risks when models generate harmful or biased content without transparent accountability mechanisms \cite{mcintosh2024inadequacy}. For instance, an LLM used in a customer service chatbot might produce inappropriate or biased responses based on hidden internal biases, without any easy way to diagnose the underlying issue or improve the LLM's behavior \cite{almeida2024exploring}. In high-stakes domains like healthcare or legal services, black-box LLMs could make critical errors without providing interpretable justifications, undermining trust and making it difficult to ensure the LLM's safety or fairness \cite{mcintosh2024cobit}. The current LLM evaluation processes could vary significantly, involving text string comparisons for straightforward answers or relying on human judgment for more complex and subtle responses \cite{chang2023survey}. 

Unlike traditional benchmarks that are often static and hardware-focused, LLM benchmarking must account for the dynamic and evolving nature of language and societal norms \cite{liang2022holistic}. LLMs can inadvertently learn and propagate biases present in training data, requiring benchmarks that can detect and mitigate such issues \cite{kenthapadi2024grounding}. Additionally, LLMs may overfit to benchmark datasets, memorizing answers rather than demonstrating true understanding, which undermines the validity of the evaluation \cite{chen2024exploring}. The potential for data contamination, where test data leaks into training datasets, further complicates the benchmarking process \cite{jegorova2022survey}. Moreover, the global deployment of LLMs requires benchmarks that consider linguistic diversity and cultural nuances, moving beyond English-centric evaluations to include multiple languages and dialects, which contrasts with traditional benchmarks that are largely language-agnostic or focused on English \cite{deng2023multilingual}. Ethical considerations are also more pronounced in LLM benchmarking, as LLMs must navigate complex social norms and avoid generating inappropriate content \cite{cheong2024not}. The aforementioned factors require a multidisciplinary approach to LLM benchmarking, integrating insights from linguistics, ethics, and social sciences to create robust evaluation frameworks \cite{chang2023survey,mcintosh2023culturally,mcintosh2024cobit}.

\section{Unified Evaluation Framework for LLM Benchmarks}
\label{sec:unified-evaluation-framework}
In this section, we developed a comprehensive framework to benchmark LLMs, grounded in principles of cybersecurity risk assessment. We based the framework on a thorough review of existing literature and best practices in the field, tailoring it to address the unique capabilities and challenges presented by generative AI especially LLMs. This framework was designed not only to assess technical performance but also to evaluate the broader applicability, robustness, and integrity of LLM benchmarks. 

\subsection{Applying the People, Process, Technology (PPT) Framework}
\label{subsec:ppt-framework-evaluation}
The People, Process, Technology (PPT) framework, widely used in cybersecurity, is also well-suited for evaluating LLM benchmarks due to its holistic approach, addressing not just technical factors, but also human and procedural elements \cite{javaid2017comprehensive}. While other frameworks (\textit{e.g.}, the Information Systems Success Model \cite{al2020evaluating}) focus on narrower aspects, PPT is uniquely comprehensive, allowing for an in-depth assessment across multiple dimensions. This ensures a thorough evaluation of benchmarks, which require more than just technical soundness; they must also integrate human expertise and follow robust, adaptable processes. The PPT framework aligns with the core challenges of LLM benchmarking. Benchmarks involve complex interactions between developers, evaluators, and end-users (\textit{People}), they must follow consistent and adaptable methodologies to remain relevant and resilient (\textit{Process}), and they depend on scalable and reliable technical infrastructures (\textit{Technology}) to accurately assess evolving LLM capabilities. The success of an LLM benchmark is not only in its ability to test specific tasks but also in how well it adapts to new challenges, handles real-world use cases, and resists manipulation or overfitting.

\begin{itemize}
	\item \textit{People}: The expertise and diversity of developers, evaluators, and end-users influence the validity and applicability of benchmark results. A lack of diversity in evaluators could bias results, making benchmarks less reflective of LLMs' real-world performance across varied scenarios.
	\item \textit{Process}: Standardization and adaptability in benchmark construction and implementation are essential. Without well-defined processes, benchmarks may become outdated, inconsistent, or vulnerable to LLMs being optimized solely for specific test sets rather than demonstrating true capability.
	\item \textit{Technology}: The technical infrastructure supporting benchmarks, including algorithms, datasets, and evaluation metrics, must be scalable and robust. Technical limitations can impede the benchmark's ability to provide accurate assessments, while ensuring the infrastructure's scalability is critical as LLMs evolve and grow more complex.
\end{itemize}

The PPT framework thus provides a holistic lens to comprehensively evaluate LLM benchmarks, addressing both functionality and integrity. By integrating people, process, and technology, this framework allows for a more nuanced assessment compared to models that may overlook key aspects of LLM benchmarking. The PPT approach aligns with the need for multidisciplinary evaluation methods in complex AI systems, as highlighted in recent literature \cite{liang2022holistic,chang2023survey}. Leveraging the PPT framework ensures our evaluation encompasses all necessary facets to produce reliable and meaningful insights into LLM performance and benchmarking practices.

\subsection{Functionality and Integrity in LLM Benchmarks}
\label{subsec:functionality-integrity-llms}
When evaluating LLM benchmarks, two core aspects must be considered: \textit{functionality} and \textit{integrity}. Functionality assesses how well a benchmark measures the specific capabilities of LLMs, while integrity examines the benchmark's resistance to manipulation or gaming by models that may exploit its criteria to achieve misleading results. Evaluating both functionality and integrity is essential to ensure that benchmarks provide a comprehensive and authentic assessment of LLMs, free from bias or manipulation.

\begin{itemize}
	\item \textit{Functionality Assessment}: This determines if the benchmark accurately evaluates the intended capabilities of an LLM, such as reasoning, language comprehension, or multimodal integration. For instance, a benchmark should be designed to test the model's true understanding rather than its ability to mimic expected answers. A counterexample is using benchmarks that rely on superficial elements like keyword matching, which can lead to overestimating an LLM's language understanding when it simply exploits patterns without grasping the content. Functionality also considers the alignment of the benchmark with real-world applications, ensuring it tests skills relevant to practical use rather than artificial metrics.
	
	\item \textit{Integrity Considerations}: This involves assessing the robustness of the benchmark against potential exploitation by models tuned to optimize for specific evaluation metrics. For example, a benchmark lacking integrity might be susceptible to LLMs that achieve high scores through memorization of training data or by exploiting specific test set characteristics. An integrity breach occurs when LLMs appear to excel in benchmarks through these surface-level optimizations without demonstrating genuine capability. Ensuring integrity means the benchmark cannot be easily manipulated and that it provides a true reflection of an LLM's performance, independent of prior exposure to similar tests or over-fitting strategies.
\end{itemize}

\subsection{Data Collection}
\label{subsec:data-collection}

We conducted a Structured Literature Review (SLR) to systematically identify relevant LLM benchmarks. Our focus was on benchmarks explicitly designed for evaluating large language models, particularly those with multimodal capabilities. The search encompassed peer-reviewed publications, prominent technical preprints, and widely cited benchmarking platforms up to October 2023.

\begin{itemize}
	\item \textbf{Inclusion Criteria:}
		\begin{itemize}
			\item Benchmarks explicitly designed for LLM evaluation.
			\item Benchmarks covering a wide range of generative AI tasks, including language understanding, reasoning, coding, legal analysis, medical question answering, and tool usage.
			\item Benchmarks with substantial academic or technical adoption, indicated by citation counts or recognition in the AI community.
		\end{itemize}
	\item \textbf{Exclusion Criteria:}
	\begin{itemize}
		\item Benchmarks that are too niche or focus on specialized tasks not representative of general LLM capabilities.
		\item Benchmarks lacking sufficient methodological detail for thorough evaluation.
	\end{itemize}
\end{itemize}

\subsection{Data Analysis}
\label{subsec:data-analysis}

We applied thematic analysis to systematically evaluate the selected benchmarks, focusing on identifying patterns of inadequacies in functionality and integrity. Our analysis aimed to determine whether the benchmarks accurately measured LLM capabilities without relying on superficial metrics or presentation nuances, and whether they could be manipulated by LLMs exploiting specific evaluation criteria. Our methodology involved the following steps:

\begin{enumerate}
	\item \textbf{Identification of Inadequacies:} We critically reviewed each benchmark to identify potential inadequacies related to functionality and integrity, such as response variability, susceptibility to overfitting, and lack of consideration for linguistic diversity.
	\item \textbf{Coding and Categorization:} We coded the identified inadequacies and categorized them based on common themes, aligning with the domains of people, process, and technology as outlined in our unified evaluation framework.
	\item \textbf{Assessment of Prevalence:} We defined \textit{prevalence} as the number of benchmarks (out of 23) exhibiting each specific inadequacy (except those resolved or not present). This metric quantifies the extent to which each inadequacy is present across the benchmarks.
	\item \textbf{Evaluation of Benchmark Responses:} For each identified inadequacy, we examined whether it was acknowledged or addressed by the benchmark creators. We categorized the benchmarks accordingly:
	\begin{itemize}
		\item \textbf{Present and Unacknowledged} (marked as \checkmark): The inadequacy exists without recognition.
		\item \textbf{Acknowledged but Unresolved} (marked as $\triangle$): The inadequacy is recognized but not yet addressed.
		\item \textbf{Considered Addressed} (marked as $\times$): The inadequacy has been recognized and mitigated.
	\end{itemize}
\end{enumerate}

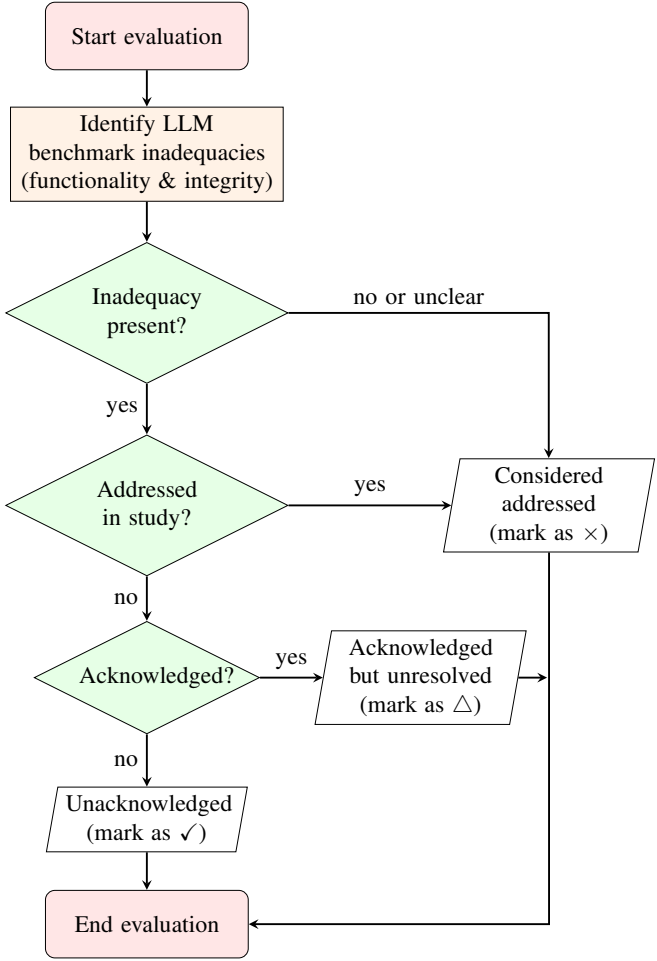
\begin{figure}[t!]
	\centering
	\resizebox{\columnwidth}{!}{
		\begin{tikzpicture}[node distance=1.35cm, auto]
			\tikzstyle{startstop} = [rectangle, rounded corners, minimum width=3cm, minimum height=1cm, text centered, draw=black, fill=red!10]
			\tikzstyle{io} = [trapezium, trapezium left angle=80, trapezium right angle=100, text centered, draw=black, text width=2.4cm] 
			\tikzstyle{process} = [rectangle, minimum width=4cm, minimum height=1cm, text centered, draw=black, fill=orange!10, text width=3.8cm]
			\tikzstyle{decision} = [diamond, aspect=2, text centered, draw=black, fill=green!10, text width=2cm]
			\tikzstyle{arrow} = [thick,->,>=stealth]
			
			\node (start) [startstop] {Start evaluation};
			\node (in1) [process, below of=start, yshift=-0.4cm] {Identify LLM benchmark inadequacies (functionality \& integrity)}; 
			\node (dec1) [decision, below of=in1, yshift=-1cm] {Inadequacy present?};
			\node (dec2) [decision, below of=dec1, yshift=-1.5cm] {Addressed in study?};
			\node (dec3) [decision, below of=dec2, yshift=-1.2cm] {Acknowledged?};
			\node (out1) [io, below of=dec3, yshift=-0.75cm] {Unacknowledged (mark as \checkmark)};
			\node (out2) [io, right of=dec2, xshift=4.6cm] {Considered addressed (mark as $\times$)};
			\node (out3) [io, right of=dec3, xshift=2.7cm] {Acknowledged but unresolved (mark as $\triangle$)};
			\node (stop) [startstop, below of=out1, yshift=-0.2cm] {End evaluation};
			
			\draw [arrow] (start) -- (in1);
			\draw [arrow] (in1) -- (dec1);
			\draw [arrow] (dec1) -- node[anchor=east] {yes} (dec2);
			\draw [arrow] (dec2) -- node[anchor=east] {no} (dec3);
			\draw [arrow] (dec2) -- node[anchor=south] {yes} (out2);
			\draw [arrow] (dec3) -- node[anchor=east] {no} (out1);
			\draw [arrow] (dec3) -- node[anchor=south] {yes} (out3);
			\draw [arrow] (out1) -- (stop);
			\draw [arrow] (dec1.east) -- node[anchor=south] {no or unclear} ++(3.85,0) -- ++(0,-2.15);
			\draw [arrow] (out2.south) |- (stop.east);
			\draw [arrow] ($(out3.east) + (0cm,0)$) -- ++(0.425cm,0); 
		\end{tikzpicture}
	}
	\caption{Evaluation Flowchart for LLM Benchmarks}
	\label{fig:Evaluation_Flowchart_for_LLM_Benchmarks}
\end{figure}

The evaluation process is illustrated in Fig.~\ref{fig:Evaluation_Flowchart_for_LLM_Benchmarks}, which guided our systematic assessment of each benchmark's robustness and reliability.

\section{Overview of LLM Benchmarks}
\label{sec:OverviewLLMBenchmarks}
In this section, we list the overview 23 LLM benchmarks we surveyed, including what they have achieved, and our preliminary findings.

\subsection{LLM Benchmarks and Their Characteristics}
Applying such criteria of our Unified Evaluation Framework, we selected 23 LLM benchmarks that represented a diverse set of tasks and domains, ensuring comprehensive coverage of the LLM evaluation landscape. Table~\ref{tab:benchmark_comparison} provides an overview of these benchmarks and their characteristics, categorized into general-purpose benchmarks and specialized-domain benchmarks.

\textbf{General-purpose benchmarks} primarily assess broad knowledge and reasoning across multiple domains. MMLU \cite{hendrycks2020measuring} from UC Berkeley focused on evaluating language understanding across diverse subjects using multiple-choice questions (MCQs) with zero-shot and few-shot evaluation methodologies. The benchmark used human exam data to test the general knowledge and reasoning abilities of LLMs in English. Chain-of-Thought Hub \cite{fu2023chain} from the University of Edinburgh used curated reasoning benchmarks to assess reasoning performance in LLMs. The benchmark featured diverse reasoning tasks in both English and Simplified Chinese, using few-shot chain-of-thought prompting and final answer accuracy as evaluation criteria. KoLA \cite{yu2023kola} from Tsinghua University focused on world knowledge evaluation using tasks related to memorization, understanding, applying, and creating knowledge. It employed standardized overall scoring and a self-contrast metric for assessing knowledge creation, using known and evolving data sources. ARB \cite{sawada2023arb} from DuckAI and Georgia Tech evaluated advanced reasoning in LLMs using standardized tests and problem books. The benchmark involved MCQs, short-answer, and open-response questions, with task-specific instructions and both automatic and manual evaluation methods. Xiezhi \cite{gu2023xiezhi} from Fudan University assessed multidisciplinary domain knowledge across 516 disciplines in English and Simplified Chinese. It used MCQs with zero-shot and few-shot evaluation, accuracy measurement, and automated test accuracy reporting. BIG-bench \cite{srivastava2022beyond} by Google focused on general knowledge, with tasks evaluated through human annotation. It used both algorithms and human raters to assess performance on diverse tasks in English. AGIEval \cite{zhong2024agieval} from Microsoft assessed human-centric reasoning tasks using official public and high-standard exams in English and Simplified Chinese. It evaluated models with MCQs and fill-in-the-blank tasks, using zero-shot and few-shot methods, chain-of-thought reasoning, and both quantitative and qualitative analysis. HELM \cite{liang2022holistic} from Stanford University compiled a benchmark collection to evaluate LLMs based on a metrics-driven approach in English. It covered 16 scenarios and 7 metrics, focusing on multi-metric measurement and dense evaluation across different tasks. PromptBench \cite{zhu2024promptbench} from Microsoft focused on general knowledge, benchmarking LLMs using a collection of APIs, datasets, and models. It employed quick performance assessments, dynamic evaluations, and semantic evaluations. C-Eval \cite{huang2023c} from Shanghai Jiao Tong University focused on multilevel, multidiscipline knowledge evaluation in Simplified Chinese using mock and high-standard exams. It employed MCQs with zero-shot and few-shot evaluation, accuracy measurement, and automated test reporting.

In contrast, \textbf{specialized benchmarks} target domain-specific LLM capabilities. HumanEval \cite{chen2021evaluating} from OpenAI was designed to assess the functional correctness of code synthesis from docstrings. It utilised programming problems with unit tests sourced from GitHub, focusing on evaluating LLMs' ability to generate accurate Python functions in English. LegalBench \cite{guha2024legalbench} by Stanford University created a benchmark for legal reasoning using manually constructed tasks. It employed IRAC (Issue, Rule, Application, Conclusion) and classification tasks, using performance evaluations and comparative analysis across models to assess their legal knowledge. FLUE \cite{shah2022flue} from Georgia Institute of Technology developed a benchmark for financial sentiment analysis and news headline classification using datasets like Financial PhraseBank and FiQA 2018. It employed sentiment analysis through regression and classification tasks to evaluate the performance of LLMs on financial data in English. MultiMedQA \cite{singhal2023large} from Google evaluated medical question answering through a collection of medical datasets such as MedQA, MedMCQA, and PubMedQA. It used MCQs, few-shot prompting, self-consistency, and human evaluation to test LLMs' performance in medical contexts. M3KE \cite{liu2023m3ke} from Tianjin University evaluated multilevel, multisubject knowledge using official exams and educational materials in Simplified Chinese. It utilised MCQs with zero-shot and few-shot evaluation methodologies, accompanied by comparative analysis across models. T-Bench \cite{xu2023tool} by SambaNova Systems Inc. focused on the tool manipulation capabilities of LLMs using empirical analysis of open-source models. It evaluated performance across diverse software tools and conducted comparative analysis of models' capabilities in real-world scenarios. SciBench \cite{wang2023scibench} from UCLA evaluated LLMs' ability to solve scientific problems using open-ended questions based on college-level textbooks and exams. It compared model outputs with correct answers and applied human-verified solutions graded by instructors' rubrics. ToolAlpaca \cite{tang2023toolalpaca} from the Chinese Academy of Sciences examined generalized tool learning in LLMs through multi-agent simulations. It evaluated performance on simulated tool-use instances across multiple categories, focusing on unseen tools and the diversity of the toolset. ToolBench \cite{qin2023toolllm} from Tsinghua University tested LLM performance in API and tool-augmented tasks. It utilized curated real-world APIs across various scenarios and assessed API retrieval accuracy, task performance, and generalization using both ground-truth and generated APIs. AgentBench \cite{liu2023agentbench} from Tsinghua University examined LLMs in agent performance across interactive tasks in English and Simplified Chinese. The benchmark involved diverse tasks in code, game, and web environments, evaluating LLMs across multiple environments through comparative analysis. APIBank \cite{li2023api} from Alibaba Group evaluated tool-augmented LLM performance in English using synthetic conversational dialogues. It measured the accuracy of API calls and response quality through the ROUGE-L metric. BOLAA \cite{liu2023bolaa} from Salesforce Research examined autonomous agent orchestration in simulated environments. It focused on decision-making and compared the performance of orchestrated agent architectures through quantitative analysis of agent interactions. HaluEval \cite{li2023halueval} from Renmin University of China investigated LLM hallucinations in English using ChatGPT-generated and human-annotated samples. The evaluation involved QA, dialogue, and summarization, relying on API-based automatic assessments of provided answers.

\begin{table*}[t]
	\centering
	\caption{Comparison of Various Generative AI and LLM Benchmarks (ranked by chronological publication dates)}
	\label{tab:benchmark_comparison}
	\resizebox{\textwidth}{!}{%
		\begin{tabular}{|l|l!{\color{gray!35}\vrule width 1pt}l!{\color{gray!35}\vrule width 1pt}l!{\color{gray!35}\vrule width 1pt}l!{\color{gray!35}\vrule width 1pt}l!{\color{gray!35}\vrule width 1pt}l|}
			\hline
			\textbf{Name} &  \textbf{Main Affiliation} & \textbf{Data Source} & \textbf{Focus Area} & \textbf{Language} & \textbf{Benchmark Format} & \textbf{Evaluation Methodology} \\ \hline
			
			MMLU \cite{hendrycks2020measuring} & UC Berkeley & Human exams & Language understanding & English & MCQs & Zero-shot and few-shot evaluation \\ \arrayrulecolor{gray!30}  \hline
			
			HumanEval \cite{chen2021evaluating} & OpenAI & GitHub code & \begin{tabular}[c]{@{}l@{}}Code Synthesis from\\Docstrings \end{tabular} & English & \begin{tabular}[c]{@{}l@{}}Programming problems\\with unit tests\end{tabular} & Functional correctness via unit tests \\
			\hline
			
			LegalBench \cite{guha2024legalbench} & \begin{tabular}[c]{@{}l@{}}Stanford\\University\end{tabular}   & \begin{tabular}[c]{@{}l@{}}Manual task\\construction\end{tabular} & Legal reasoning & English & \begin{tabular}[c]{@{}l@{}}IRAC \& classification\\tasks\end{tabular}  & \begin{tabular}[c]{@{}l@{}}Task performance evaluation,\\comparative analysis across models\end{tabular} \\ \hline
			
			FLUE \cite{shah2022flue} & \begin{tabular}[c]{@{}l@{}}Georgia Institute\\of Technology\end{tabular}  & \begin{tabular}[c]{@{}l@{}}Financial PhraseBank,\\ FiQA 2018, Gold news\\headline dataset\end{tabular} & \begin{tabular}[c]{@{}l@{}}Financial sentiment\\ analysis, news\\headline classification\end{tabular} & English & \begin{tabular}[c]{@{}l@{}}Sentiment analysis\\(regression, classification)\end{tabular} & \begin{tabular}[c]{@{}l@{}}Evaluation on datasets\end{tabular} \\ \hline

			MultiMedQA \cite{singhal2023large} & Google & \begin{tabular}[c]{@{}l@{}}MedQA, MedMCQA,\\PubMedQA, \textit{etc.}\end{tabular} & \begin{tabular}[c]{@{}l@{}}Medical question\\answering\end{tabular} & English & MCQs & \begin{tabular}[c]{@{}l@{}}Few-shot prompting, self\\-consistency, human evaluation\end{tabular} \\
			\hline

			M3KE \cite{liu2023m3ke} & \begin{tabular}[c]{@{}l@{}}Tianjin\\University\end{tabular}  & \begin{tabular}[c]{@{}l@{}}Official exams and\\educational materials\end{tabular} & \begin{tabular}[c]{@{}l@{}}Multilevel, multisubject\\knowledge evaluation\end{tabular} & Simplified Chinese & MCQs & \begin{tabular}[c]{@{}l@{}}Zero-shot and few-shot evaluation,\\comparative analysis across models\end{tabular} \\ \hline
			
			T-Bench \cite{xu2023tool} & \begin{tabular}[c]{@{}l@{}}SambaNova\\Systems Inc.\end{tabular}  & \begin{tabular}[c]{@{}l@{}}Empirical analysis of\\open-source LLMs\end{tabular}
			&\begin{tabular}[c]{@{}l@{}}Software tool\\manipulation capability\end{tabular} & English & Diverse software tools &
			\begin{tabular}[c]{@{}l@{}}Comparative analysis across models,\\performance evaluation \end{tabular} \\ \hline

			\begin{tabular}[c]{@{}l@{}}Chain-of-Thought\\Hub \cite{fu2023chain}\end{tabular} & \begin{tabular}[c]{@{}l@{}}University of\\Edinburgh\end{tabular} & \begin{tabular}[c]{@{}l@{}}Curated reasoning\\benchmarks\end{tabular} & Reasoning performance & \begin{tabular}[c]{@{}l@{}}English,\\Simplified Chinese\end{tabular}  & Diverse reasoning tasks & \begin{tabular}[c]{@{}l@{}}Few-shot chain-of-thought prompting,\\Final answer accuracy evaluation\end{tabular} \\ \hline

			KoLA \cite{yu2023kola} & \begin{tabular}[c]{@{}l@{}}Tsinghua\\University\end{tabular} & \begin{tabular}[c]{@{}l@{}}Known and evolving\\data sources \end{tabular} & World knowledge & English & \begin{tabular}[c]{@{}l@{}}Knowledge memorization, \\understanding, applying, \\and creating tasks\end{tabular} & \begin{tabular}[c]{@{}l@{}}Standardized overall scoring, \\self-contrast metric for \\knowledge creation\end{tabular} \\ \hline

			SciBench \cite{wang2023scibench} & UC LA & \begin{tabular}[c]{@{}l@{}}College-level\\textbooks and\\course exams\end{tabular} & Scientific problems & English & Open-ended questions & \begin{tabular}[c]{@{}l@{}}Comparison with correct answers;\\graded by instructors' rubrics;\\human-verified solutions\end{tabular}
			\\ \hline

			ARB \cite{sawada2023arb} & \begin{tabular}[c]{@{}l@{}}DuckAI \&\\Georgia Tech\end{tabular} & \begin{tabular}[c]{@{}l@{}}Standardized tests, \\ problem books\end{tabular} & Advanced reasoning & English & \begin{tabular}[c]{@{}l@{}}MCQs, short answer,\\open response questions\end{tabular} & \begin{tabular}[c]{@{}l@{}}Task-specific instructions, \\automatic and manual evaluation\end{tabular} \\ \hline

			Xiezhi \cite{gu2023xiezhi} & Fudan University & \begin{tabular}[c]{@{}l@{}}Comprehensive domain\\knowledge evaluation\end{tabular} & \begin{tabular}[c]{@{}l@{}}Multidisciplinary,\\516 disciplines\end{tabular} & \begin{tabular}[c]{@{}l@{}}English,\\Simplified Chinese\end{tabular} & MCQs & \begin{tabular}[c]{@{}l@{}}Zero-shot and few-shot evaluation,\\accuracy measurement,\\automated test accuracy reporting\end{tabular} \\ \hline

			BIG-bench \cite{srivastava2022beyond}  & Google & Human annotation & General knowledge & English & Tasks & Algorithm and human raters \\ \hline
			
			AGIEval \cite{zhong2024agieval} & Microsoft & \begin{tabular}[c]{@{}l@{}}Official public and\\high-standard exams\end{tabular} & \begin{tabular}[c]{@{}l@{}}Human-centric reasoning\\tasks\end{tabular} & \begin{tabular}[c]{@{}l@{}}English,\\Simplified Chinese\end{tabular} & \begin{tabular}[c]{@{}l@{}}MCQs,\\ fill-in-the-blank\end{tabular} & \begin{tabular}[c]{@{}l@{}}Zero-shot and few-shot evaluation,\\ chain-of-thought reasoning,\\ quantitative and qualitative analysis\end{tabular} \\ \hline

			ToolAlpaca \cite{tang2023toolalpaca} & \begin{tabular}[c]{@{}l@{}}Chinese\\Academy\\ of Sciences\end{tabular} & Multi-agent simulation & \begin{tabular}[c]{@{}l@{}}Generalized tool learning\\for language models\end{tabular} & English & \begin{tabular}[c]{@{}l@{}}Simulated tool-use\\instances in\\ multiple categories\end{tabular} & \begin{tabular}[c]{@{}l@{}}Performance evaluation on\\unseen tools, diversity in toolset\end{tabular} \\ \hline
			
			HELM \cite{liang2022holistic} & \begin{tabular}[c]{@{}l@{}}Stanford\\University\end{tabular} & Benchmark collection & Metrics-based evaluation & English & 16 Scenarios, 7 metrics & \begin{tabular}[c]{@{}l@{}}Multi-metric measurement, Dense\\ evaluation of scenarios and metrics\end{tabular} \\ \hline
			
			ToolBench \cite{qin2023toolllm} & \begin{tabular}[c]{@{}l@{}}Tsinghua\\University\end{tabular}   & \begin{tabular}[c]{@{}l@{}}Curated from 16,000+\\real-world APIs using\\ChatGPT\end{tabular} & \begin{tabular}[c]{@{}l@{}}API and tool-augmented\\LLM performance\end{tabular} & English & \begin{tabular}[c]{@{}l@{}}API-based tasks in\\various scenarios\end{tabular} & \begin{tabular}[c]{@{}l@{}}Assessment of API retrieval accuracy,\\task performance, and generalization\\ capability using both ground-truth\\and generated APIs\end{tabular} \\ \hline

			PromptBench \cite{zhu2024promptbench}  & Microsoft & Benchmark collection & General knowledge & English & APIs, datasets, models & \begin{tabular}[c]{@{}l@{}}Quick performance assessment,\\dynamic evaluation, semantic\\evaluation\end{tabular} \\ \hline

			AgentBench \cite{liu2023agentbench} & \begin{tabular}[c]{@{}l@{}}Tsinghua\\University\end{tabular}  & \begin{tabular}[c]{@{}l@{}}Code, game, and web\\ environments\end{tabular} & \begin{tabular}[c]{@{}l@{}}Agent performance \\in interactive tasks\end{tabular} & \begin{tabular}[c]{@{}l@{}}English, \\ Simplified Chinese\end{tabular} & \begin{tabular}[c]{@{}l@{}}Diverse interactive tasks\\across environments\end{tabular} & \begin{tabular}[c]{@{}l@{}}Evaluation across multiple environments,\\comparative analysis of LLMs as agents\end{tabular} \\ \hline

			APIBank \cite{li2023api} & Alibaba Group & Synthetic dialogues  &
			\begin{tabular}[c]{@{}l@{}}Tool-augmented LLM\\performance\end{tabular} & English & Conversational dialogues & 			\begin{tabular}[c]{@{}l@{}}Accuracy of API calls and response\\quality measured by ROUGE-L metric \end{tabular}
			\\ \hline

			C-Eval \cite{huang2023c} & \begin{tabular}[c]{@{}l@{}}Shanghai \\Jiao Tong\\University\end{tabular} & \begin{tabular}[c]{@{}l@{}}Mock exams, high\\-standard exams\end{tabular} & \begin{tabular}[c]{@{}l@{}}Multilevel, multidiscipline \\ knowledge evaluation\end{tabular} & Simplified Chinese & MCQs & \begin{tabular}[c]{@{}l@{}}Zero-shot and few-shot evaluation, \\ Accuracy measurement, \\ Automated test accuracy reporting\end{tabular} \\ \hline
			
			BOLAA \cite{liu2023bolaa} & \begin{tabular}[c]{@{}l@{}}Salesforce\\Research (USA)\end{tabular}  & \begin{tabular}[c]{@{}l@{}}Simulated\\environments\end{tabular} & \begin{tabular}[c]{@{}l@{}}Autonomous agent \\orchestration \end{tabular}  & English & Decision-making & \begin{tabular}[c]{@{}l@{}}Performance comparison of orchestrated\\agent architectures, quantitative\\analysis of agent interactions\end{tabular} \\ \hline

			HaluEval \cite{li2023halueval} & \begin{tabular}[c]{@{}l@{}}Renmin\\ University\\of China \end{tabular} & \begin{tabular}[c]{@{}l@{}}ChatGPT-generated\\ and human-annotated\\samples \end{tabular}
			& LLM hallucinations & English & \begin{tabular}[c]{@{}l@{}}QA, dialogue, and\\summarization\end{tabular} & \begin{tabular}[c]{@{}l@{}}API-based automatic evaluation\\according to provided answers\end{tabular} \\ \arrayrulecolor{black}  \hline
			
		\end{tabular}%
	}
\end{table*}

\subsection{Preliminary Findings Overview}
\label{subsec:preliminary-findings}

Our analysis, as summarised in Table \ref{tab:BenchmarkComparison}, revealed widespread inadequacies across LLM benchmark studies, indicating the need for more refined and comprehensive evaluation practices to better assess the true capabilities and limitations of large language models (LLMs). A key observation is that many benchmarks predominantly focus on English, with limited representation of other languages, such as Simplified Chinese, and frequently neglect cultural and contextual variations in their questions and answers. This raises concerns about the generalizability of these benchmarks across diverse linguistic and cultural settings. For instance, benchmarks often assume singular correct answers in culturally nuanced questions, overlooking alternative valid responses from different cultural or religious perspectives, which limits the inclusivity of these evaluations. Another critical issue is that current benchmarks generally adopt task-based formats, such as Multiple Choice Questions (MCQs) and dialogue-based evaluations, which tend to be static and do not capture the evolving nature of human-AI interactions. Real-world LLM usage often involves continuous dialogues, yet many benchmarks assess only the first attempt of an LLM response, without considering the consistency and coherence of answers across multiple interactions. This focus on isolated responses reduces the relevance of these benchmarks in evaluating models designed for dynamic, ongoing interactions.

Our review also found that only 6 of the 23 surveyed benchmark studies (\cite{hendrycks2020measuring,liang2022holistic,li2023halueval,shah2022flue,singhal2023large,zhu2024promptbench}) were peer-reviewed at the time of this article, reflecting the early-stage nature of research in this domain. While preprints provide valuable insights, the lack of rigorous peer review raises questions about the scientific validity and reproducibility of many LLM benchmark results. Moreover, the speed of LLM output generation—a crucial factor for user experience in real-time applications—is frequently overlooked in current benchmarks, which tend to focus solely on the qualitative correctness of generated answers. Perhaps most concerning, many benchmarks failed to account for the possibility that LLMs can optimize their responses specifically to perform well on standardized tests, rather than genuinely demonstrating deep understanding or reasoning. This risk of "benchmark gaming" undermines the integrity of evaluations, as models may be engineered to exploit the structure of the test rather than showcasing their full capabilities across diverse tasks. Given the rapid advancement of LLMs, it is imperative to develop evaluation practices that measure genuine reasoning skills and adaptability, rather than technical optimization for specific test formats. Our critique is grounded in published opinions and supported by our evaluative rationale, detailed in Sections \ref{sec:technological-aspects}, \ref{sec:processual-elements}, and \ref{sec:human-dynamics}. The aim of this analysis is not to elevate any specific benchmark but to highlight common limitations that many benchmarks share. By addressing these issues, we seek to encourage the development of more robust, transparent, and representative benchmarks that better reflect the full range of LLM capabilities. The following sections provide an in-depth exploration of these findings, categorized into technological aspects, processual elements, and human dynamics.

\begin{table*}[!t]
	\centering
	\caption{Prevalence of LLM Benchmark Inadequacies\\ (\checkmark: present and unacknowledged; $\triangle$: acknowledged but unresolved; $\times$: resolved or not present)}
	\label{tab:BenchmarkComparison}
	\resizebox{\textwidth}{!}{%
		\begin{tabular}{|c|c!{\color{gray!35}\vrule width 1pt}c!{\color{gray!35}\vrule width 1pt}c!{\color{gray!35}\vrule width 1pt}c!{\color{gray!35}\vrule width 1pt}c!{\color{gray!35}\vrule width 1pt}c|c!{\color{gray!35}\vrule width 1pt}c!{\color{gray!35}\vrule width 1pt}c|c!{\color{gray!35}\vrule width 1pt}c|}
			\hline
			\multirow{2}{*}{\begin{tabular}[c]{@{}c@{}}~\\~\\~\\Benchmark\\ Research\end{tabular}} & \multicolumn{6}{c|}{Technological Aspects (Sec \ref{sec:technological-aspects})} & \multicolumn{3}{c|}{Processual Elements (Sec \ref{sec:processual-elements})} & \multicolumn{2}{c|}{Human Dynamics (Sec \ref{sec:human-dynamics})} \\ \cline{2-12} 
			& \multicolumn{1}{c!{\color{gray!35}\vrule width 1pt}}{\rotatebox[origin=c]{90}{\begin{tabular}[c]{@{}c@{}}Response Variability in\\Standardized Evaluations (Sec \ref{subsec:response-variability-standardized-evaluations})\end{tabular}}} & \multicolumn{1}{c!{\color{gray!35}\vrule width 1pt}}{\rotatebox[origin=c]{90}{\begin{tabular}[c]{@{}c@{}}Genuine Reasoning vs Technical\\Optimization (Sec \ref{subsec:genuine-reasoning-optimization})\end{tabular}}} & \multicolumn{1}{c!{\color{gray!35}\vrule width 1pt}}{\rotatebox[origin=c]{90}{\begin{tabular}[c]{@{}c@{}}~Tension Between Helpfulness~\\and Harmlessness (sec:\ref{subsec:tension-helpfulness-harmlessness})\end{tabular}}} & \multicolumn{1}{c!{\color{gray!35}\vrule width 1pt}}{\rotatebox[origin=c]{90}{\begin{tabular}[c]{@{}c@{}}Linguistic Variability and\\Embedded Logic Diversity (Sec \ref{subsec:linguistic-differences-embedded-logics}) \end{tabular}}} & \multicolumn{1}{c!{\color{gray!35}\vrule width 1pt}}{\rotatebox[origin=c]{90}{\begin{tabular}[c]{@{}c@{}}Benchmark Installation\\and Scalability (Sec \ref{subsec:benchmark-installation-scalability})\end{tabular}}} & \rotatebox[origin=c]{90}{\begin{tabular}[c]{@{}c@{}}Biases in LLM-Generated \\LLM Evaluations (Sec \ref{subsec:biases-LLM-generated-evaluations})\end{tabular}} & \multicolumn{1}{c!{\color{gray!35}\vrule width 1pt}}{\rotatebox[origin=c]{90}{\begin{tabular}[c]{@{}c@{}}Inconsistent Benchmark\\Implementation (Sec \ref{subsec:inconsistent-benchmark-implementation-inconsistency})\end{tabular}}} & \multicolumn{1}{c!{\color{gray!35}\vrule width 1pt}}{\rotatebox[origin=c]{90}{\begin{tabular}[c]{@{}c@{}}Slow Test Iteration time (Sec \ref{subsec:slow-iteration-time})\end{tabular}}} & \rotatebox[origin=c]{90}{\begin{tabular}[c]{@{}c@{}}Challenge of Proper Prompt\\Engineering (Sec \ref{subsec:prompt-engineering-challenge})\end{tabular}} & \multicolumn{1}{c!{\color{gray!35}\vrule width 1pt}}{\rotatebox[origin=c]{90}{\begin{tabular}[c]{@{}c@{}}Diversity in Human Curators\\and Evaluators (Sec \ref{subsec:diversity-human-curators-evaluators})\end{tabular}}} & \rotatebox[origin=c]{90}{\begin{tabular}[c]{@{}c@{}}Diverse Cultural, Social,\\Political, Religious and\\Ideological Norms (Sec \ref{subsec:diverse-cultural-ideological-challenges})\end{tabular}} \\ \hline
			
			MMLU \cite{hendrycks2020measuring} & \checkmark & \checkmark & $\triangle$ & $\triangle$ & \checkmark  & $\times$ & \checkmark & \checkmark & $\triangle$ & $\triangle$ & $\triangle$ \\ \arrayrulecolor{gray!30}  \hline
			
			HumanEval \cite{chen2021evaluating} &  \checkmark & \checkmark  & \checkmark  & \checkmark  & $\times$  &  $\times$ & $\triangle$ & $\times$ & $\times$ & \checkmark & $\times$ \\ \hline
			
			LegalBench \cite{guha2024legalbench} &  \checkmark &  \checkmark & \checkmark  & \checkmark & \checkmark & $\times$ & $\triangle$ & $\times$ & $\triangle$ & $\triangle$ & \checkmark \\ \hline
			
			FLUE \cite{shah2022flue} & \checkmark & \checkmark & \checkmark  & \checkmark  &  $\times$ & $\times$  & \checkmark  &  \checkmark & \checkmark &  $\times$ & \checkmark \\ \hline
			
			MultiMedQA \cite{singhal2023large} &  \checkmark & $\triangle$  & \checkmark  & $\triangle$  &  $\times$ &  $\times$ &  $\triangle$ & \checkmark  & $\triangle$ & $\triangle$  & $\triangle$ \\ \hline
			
			M3KE \cite{liu2023m3ke} &  \checkmark &  \checkmark & \checkmark  & \checkmark  & \checkmark  &  \checkmark & \checkmark &  $\times$ & \checkmark  &  \checkmark & $\triangle$ \\ \hline
			
			T-Bench \cite{xu2023tool} & \checkmark  &  \checkmark & \checkmark  &  \checkmark &  \checkmark & \checkmark  & $\times$ &  $\triangle$  & $\triangle$ & \checkmark  & $\times$ \\ \hline
			
			Chain-of-Thought Hub \cite{fu2023chain} & $\triangle$  &  \checkmark &  \checkmark & \checkmark  & $\times$  & $\times$  & \checkmark &  \checkmark  &  $\triangle$ & \checkmark  & \checkmark \\ \hline
			
			KoLA \cite{yu2023kola} & \checkmark & $\times$ & \checkmark & $\triangle$ & \checkmark &  $\times$ & $\triangle$ & \checkmark & $\times$ & \checkmark  & $\triangle$ \\ \hline
			
			SciBench \cite{wang2023scibench} &  \checkmark & \checkmark  &  \checkmark &  $\times$ & \checkmark  & $\times$  & $\triangle$ & $\triangle$  & $\times$ & $\triangle$ & \checkmark \\ \hline
			
			ARB \cite{sawada2023arb} &  $\triangle$ &  \checkmark & \checkmark  &  $\times$ & \checkmark  & \checkmark  & $\triangle$ & $\triangle$  & $\triangle$ & $\times$  & \checkmark \\ \hline
			
			Xiezhi \cite{gu2023xiezhi} & $\times$  & \checkmark  & $\times$  & \checkmark  &  \checkmark &  \checkmark & \checkmark  & $\times$  & $\times$ &  $\triangle$ & $\triangle$ \\ \hline
			
			BIG-Bench \cite{srivastava2022beyond} & \checkmark  & \checkmark  & $\times$  & $\times$  & \checkmark  &  \checkmark & $\times$ &  \checkmark & $\triangle$ &  $\triangle$ & \checkmark \\ \hline
			
			AGIEval \cite{zhong2024agieval} & \checkmark  &  \checkmark & \checkmark  & $\triangle$  & \checkmark  & $\times$  & \checkmark  &  \checkmark &   $\times$  & $\triangle$  & \checkmark \\ \hline
			
			ToolAlpaca \cite{tang2023toolalpaca} & \checkmark & \checkmark & $\times$ & \checkmark & \checkmark  & \checkmark & \checkmark & \checkmark  & $\times$ & $\triangle$  & $\times$ \\ \hline
			
			HELM \cite{liang2022holistic} &  $\triangle$ & \checkmark  &  \checkmark & $\triangle$  &  $\times$  & $\times$   & $\triangle$ & $\triangle$  & $\triangle$ & $\times$   & $\triangle$  \\ \hline
			
			ToolBench \cite{qin2023toolllm} & \checkmark  & \checkmark  &  \checkmark & \checkmark  &  $\times$  &  $\times$  & $\triangle$ &  $\times$  & $\times$  &  \checkmark & $\times$  \\ \hline
			
			PromptBench \cite{zhu2024promptbench} &  \checkmark & $\triangle$  & \checkmark  & $\times$   & \checkmark  &  $\times$  & \checkmark &  \checkmark &   $\times$   & \checkmark  & \checkmark \\ \hline
			
			AgentBench \cite{liu2023agentbench} & \checkmark  & \checkmark  & \checkmark  &  $\times$  &  \checkmark & \checkmark  & \checkmark &  \checkmark & $\triangle$ &  \checkmark & \checkmark \\ \hline
			
			APIBank \cite{li2023api} &  \checkmark &  \checkmark & $\times$  &  $\triangle$ &  \checkmark & \checkmark  & $\times$  & \checkmark  & $\times$  &  \checkmark & \checkmark \\ \hline
			
			C-Eval \cite{huang2023c} & $\triangle$  & \checkmark  &  \checkmark & \checkmark  & $\times$   & $\times$   & $\times$  & \checkmark  & \checkmark &  \checkmark & $\triangle$ \\ \hline
			
			BOLAA \cite{liu2023bolaa} & \checkmark  & $\times$  & $\times$  &  $\times$  &  $\times$ & $\times$   & $\times$  & $\times$  & $\triangle$ &  $\times$  & $\times$ \\ \hline
			
			HaluEval \cite{li2023halueval} &  $\times$ & $\times$  & $\times$  & $\times$  &  $\times$ &  $\times$ & $\triangle$ & $\triangle$ & $\triangle$ &  $\triangle$ & $\times$ \\  \arrayrulecolor{black}  \hline
		\end{tabular}%
	}
\end{table*}

\section{Technological Aspects}
\label{sec:technological-aspects}
\begin{figure}[t!]
	\centering
	\resizebox{\columnwidth}{!}{
		\begin{tikzpicture}
			\begin{axis}[
				xbar, xmin=0, xmax=23,
				width=0.7\columnwidth,
				height=0.85\columnwidth,
				enlarge y limits=0.1,
				xlabel={Prevalence (max 23)},
				symbolic y coords={
					{Biases in LLM\\-Generated Evaluations},
					{Benchmark Installation\\and Scalability},
					{Linguistic Variability and\\Embedded Logic Diversity},
					{Tension - Helpfulness\\and Harmlessness},
					{Genuine Reasoning vs\\Technical Optimization},
					{Response Variability in\\Standardized Evaluations}
				},
				ytick=data,
				yticklabel style={align=center},
				nodes near coords, nodes near coords align={horizontal},
				every node near coord/.append style={
					anchor=center, 
					xshift=-0.5cm, 
					color=black
				},
				]
				\addplot[fill=cyan!10]coordinates {
					(9,{Biases in LLM\\-Generated Evaluations})
					(16,{Benchmark Installation\\and Scalability})
					(17,{Linguistic Variability and\\Embedded Logic Diversity})
					(19,{Tension - Helpfulness\\and Harmlessness})
					(22,{Genuine Reasoning vs\\Technical Optimization})
					(22,{Response Variability in\\Standardized Evaluations})
				};
			\end{axis}
		\end{tikzpicture}
	}
	\caption{Technological Inadequacies in LLM Benchmarking}
	\label{fig:Technological_Inadequacies}
\end{figure}
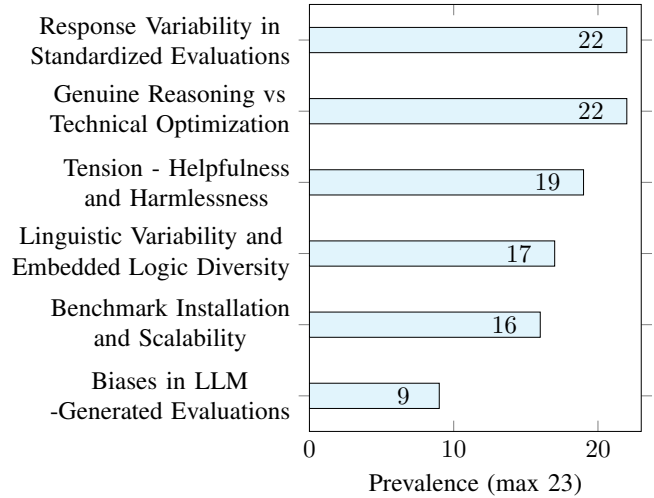

In this section, we examine the technological intricacies and limitations inherent in current LLM benchmarks (Fig. \ref{fig:Technological_Inadequacies}).

\subsection{Response Variability in Standardized Evaluations}
\label{subsec:response-variability-standardized-evaluations}
\textbf{Prevalence}: 22/23

A significant inadequacy in LLM benchmarks is the response variability under standardized evaluations, especially when these benchmarks fail to account for LLMs tailored to specific formats or use cases. Despite the breadth of frameworks like \cite{chen2021evaluating,fu2023chain,guha2024legalbench,hendrycks2020measuring,huang2023c,liang2022holistic,liu2023agentbench,li2023api,li2023halueval,liu2023bolaa,liu2023m3ke,qin2023toolllm,sawada2023arb,shah2022flue,singhal2023large,srivastava2022beyond,tang2023toolalpaca,wang2023scibench,xu2023tool,yu2023kola,zhong2024agieval,zhu2024promptbench}, they often overlook the subtle behaviors of LLMs designed for particular scenarios (Appendix \ref{subsec:Appendix1-response-variability-standardized-evaluations}). Benchmarks that standardize formats without considering context-specific requirements can inadvertently skew the perceived functionality of these LLMs. For example, minor formatting changes in prompts, such as altering choice indicators from (A) to [A] or adding extra spaces, have been shown to shift response accuracy by approximately 5\%, highlighting the LLMs' sensitivity to superficial input variations \cite{ganguli2023challenges}. From a functionality perspective, this variability raises concerns about the accuracy of these benchmarks in measuring true model capabilities. If benchmarks fail to reflect the intended application context, they may mislead users and developers regarding the model's practical performance, influencing deployment strategies inappropriately \cite{chang2023survey}. Integrity concerns also arise when standardized benchmarks become predictable and exploitable, allowing LLMs to achieve artificially high scores through superficial optimization rather than genuine comprehension. This vulnerability to input sensitivity indicates a lack of robustness in the benchmarks, which could be exploited in practical applications, posing risks to system reliability and security \cite{wei2023jailbroken}. To address such issues, LLM benchmark designs must be refined to accommodate the specificities of model architecture and intended use cases, ensuring they accurately capture the capabilities of LLMs in varied contexts. This approach will enhance the functionality and integrity of the evaluations, reducing the potential for misrepresentation and gaming of results, and ultimately leading to a more authentic assessment of LLM capabilities.

\subsection{Genuine Reasoning vs Technical Optimization}
\label{subsec:genuine-reasoning-optimization}
\textbf{Prevalence}: 22/23

The challenge in distinguishing between responses driven by genuine reasoning and those resulting from technical optimization, such as overfitting to benchmark answers, was seen in \cite{chen2021evaluating,fu2023chain,gu2023xiezhi,guha2024legalbench,hendrycks2020measuring,huang2023c,li2023api,li2023halueval,liang2022holistic,liu2023agentbench,liu2023bolaa,liu2023m3ke,qin2023toolllm,sawada2023arb,shah2022flue,singhal2023large,srivastava2022beyond,tang2023toolalpaca,wang2023scibench,xu2023tool,zhong2024agieval,zhu2024promptbench} (Appendix \ref{subsec:Appendix1-discerning-reasoning-optimization}), due to the opaque nature of LLMs, where the mechanisms behind their outputs are often not transparent. Consequently, LLMs can appear to demonstrate advanced reasoning when, in reality, they may simply be leveraging specific benchmark characteristics or exploiting superficial patterns. For example, an LLM trained on data that includes benchmark-like questions may produce correct answers by recognizing patterns rather than understanding the underlying concepts \cite{balloccu2024leak,ganguli2023challenges}. For example, HumanEval \cite{chen2021evaluating} evaluated Codex's performance through functional correctness on programming problems, which might not fully capture the model's reasoning capabilities across diverse real-world scenarios. Similarly, LegalBench \cite{guha2024legalbench} applied generic benchmarks to specialized legal reasoning tasks, potentially allowing LLMs to optimize for benchmark-specific patterns rather than demonstrating true legal comprehension that could include court negotiations.

From a functionality perspective, this ambiguity undermines the benchmarks' ability to accurately assess LLMs' reasoning capabilities. If a benchmark fails to differentiate between true reasoning and optimization, it risks overestimating the model's practical utility and cognitive abilities. A critical insight here is that benchmarks should be designed to probe deeper understanding rather than surface-level pattern recognition, ensuring that LLMs cannot succeed merely by exploiting familiar test elements. Regarding integrity, the inability to distinguish genuine reasoning from optimization raises concerns about the potential for LLMs to manipulate benchmark outcomes. LLMs that can achieve high scores through memorization or strategic over-fitting challenge the validity of the evaluation process. This issue is particularly problematic when benchmarks are inadvertently included in training data, allowing LLMs to produce artificially inflated results without demonstrating true competence \cite{balloccu2024leak,zhou2023don}. Addressing this requires designing benchmarks that are resistant to such exploitation, ensuring they accurately reflect the LLM's ability to engage in authentic reasoning processes rather than optimized mimicry. To mitigate such challenges, benchmark designs must evolve to include tasks that require dynamic reasoning and adaptability, reducing the risk of LLMs gaming the system through technical optimizations. Continuous refinement and innovation in benchmark methodologies are essential to differentiate genuine reasoning from mere pattern exploitation, thereby providing a more reliable assessment of LLM capabilities.

\subsection{Tension Between Helpfulness and Harmlessness}
\label{subsec:tension-helpfulness-harmlessness}
\textbf{Prevalence}: 19/23

The tension between helpfulness and harmlessness in LLM responses presents a significant challenge in benchmark evaluations, seen in \cite{chen2021evaluating,fu2023chain,guha2024legalbench,hendrycks2020measuring,huang2023c,li2023halueval,liang2022holistic,liu2023agentbench,liu2023bolaa,liu2023m3ke,qin2023toolllm,sawada2023arb,shah2022flue,singhal2023large,wang2023scibench,xu2023tool,yu2023kola,zhong2024agieval,zhu2024promptbench} (Appendix \ref{subsec:Appendix1-tension-helpfulness-harmlessness}). Human evaluations, such as A/B tests, often struggle to balance those two aspects, particularly in open-ended dialogues. For instance, MultiMedQA \cite{singhal2023large} assessed Flan-PaLM's performance in medical question answering, revealing that while the model could provide accurate information, it occasionally failed to align with clinical consensus, indicating a struggle to balance helpfulness with the necessity of harmlessness in sensitive medical contexts. Similarly, LegalBench \cite{guha2024legalbench} evaluated LLMs on legal reasoning tasks, where the models needed to deliver precise legal information without overstepping into providing authoritative legal advice, thereby maintaining a balance between being helpful and avoiding potential legal ramifications. Such difficulty arises from the need to provide information without causing harm or controversy, a balance that is particularly delicate when addressing sensitive topics.

From a functionality perspective, LLM benchmarks must accurately measure an LLM's ability to provide informative yet harmless responses. This requires sophisticated evaluation criteria that distinguish between helpfulness in providing useful information and harmlessness in avoiding misinformation or offensive content. Benchmarks that fail to account for this balance risk either over-penalizing LLMs for providing valuable but sensitive information or underestimating the potential for harm in their responses \cite{chang2023survey}. An insightful example is the evaluation of LLMs in the context of cybersecurity advice, where helpfulness could involve explaining malware analysis techniques, but doing so without risking the dissemination of potentially harmful information. Regarding integrity, benchmarks must be robust against LLMs that could exploit this tension to appear more capable \cite{mcintosh2024inadequacy}. For instance, LLMs might evade challenging inquiries by refusing to respond, prioritizing harmlessness over helpfulness to avoid any potential controversy. This behavior could lead to a misleading assessment of the model's abilities, giving the impression of ethical consideration when it may simply be avoiding complexity. To ensure integrity, benchmarks should test for situations where LLMs can demonstrate both helpfulness and harmlessness without compromising one for the other. Addressing this inadequacy requires developing benchmarks that incorporate high-level norms and values, ensuring a comprehensive evaluation of how LLMs navigate this tension.

\subsection{Linguistic Variability and Embedded Logic Diversity}
\label{subsec:linguistic-differences-embedded-logics}
\textbf{Prevalence}: 17/23

Current LLM benchmarks often fail to account for the linguistic and logical diversity inherent in different languages, leading to an inadequate evaluation of LLMs' true capabilities, as seen in \cite{chen2021evaluating,fu2023chain,gu2023xiezhi,guha2024legalbench,hendrycks2020measuring,huang2023c,li2023api,li2023halueval,liang2022holistic,liu2023m3ke,qin2023toolllm,shah2022flue,singhal2023large,tang2023toolalpaca,xu2023tool,yu2023kola,zhong2024agieval} (Appendix \ref{subsec:Appendix1-linguistic-differences-embedded-logics}). Predominantly English-centric benchmarks overlook the sophisticated reasoning patterns embedded within languages, when language structure can shape distinct cognitive processes \cite{perlovsky2009language,mcintosh2024inadequacy}. This lack of consideration results in benchmarks that wrongfully assume a uniform cognitive framework, failing to capture how LLMs handle linguistic variations across different cultures and languages \cite{mcintosh2023culturally}.

From a functionality perspective, this inadequacy undermines the benchmarks' ability to evaluate LLMs in a multilingual context accurately. By not reflecting the cognitive diversity inherent in language structures, these benchmarks may misrepresent a model's comprehension and reasoning capabilities. For instance, an LLM evaluated solely on English-centric benchmarks might appear proficient but could struggle to maintain accuracy and context in languages with different syntactic and semantic frameworks. This bias risks overestimating a model's universality and applicability across languages. In terms of integrity, the oversight in linguistic variability can be exploited, as uneven moderation across languages may allow harmful or prohibited content to bypass detection in less moderated languages \cite{deng2023multilingual}. A model might score well on benchmarks in its primary language but fail to uphold the same standards in others, potentially leading to security vulnerabilities. Besides, a blend of different languages could sometimes be used to jailbreak LLMs without triggering LLM safety mechanisms like RLHF \cite{mcintosh2024inadequacy}. Addressing such challenges requires benchmarks that recognize and adapt to linguistic diversity, testing LLMs on native language capabilities rather than relying on translation-based evaluations.

\subsection{Benchmark Installation and Scalability}
\label{subsec:benchmark-installation-scalability}
\textbf{Prevalence}: 16/23

A key inadequacy in current LLM benchmarks is the challenge of installation and scalability, affecting their functional utility and integrity \cite{gu2023xiezhi,guha2024legalbench,hendrycks2020measuring,li2023api,li2023halueval,liu2023agentbench,liu2023bolaa,liu2023m3ke,sawada2023arb,srivastava2022beyond,tang2023toolalpaca,wang2023scibench,xu2023tool,yu2023kola,zhong2024agieval,zhu2024promptbench} (Appendix \ref{subsec:Appendix1-benchmark-installation-scalability}). Many benchmarks demand considerable engineering effort to install and adapt to different computational environments, hindering broad accessibility and consistent evaluation across diverse LLMs. For example, SciBench \cite{wang2023scibench} involved manual data extraction and complex problem formulations, further complicating the installation process. Moreover, scaling such benchmarks to handle larger LLMs or more extensive datasets often requires substantial infrastructure and resources, as reported by Anthropic \cite{ganguli2023challenges}, posing a barrier to comprehensive and efficient assessments. For instance, APIBank \cite{li2023api} involved the complex implementation of 73 APIs, demanding extensive engineering efforts and resources to scale effectively.

From a functionality standpoint, these technical hurdles delay and complicate the evaluation process, potentially limiting the ability to accurately measure LLM performance. For instance, if a benchmark cannot be efficiently scaled, it may fail to test an LLM's capabilities under more demanding or realistic scenarios, resulting in an incomplete functional assessment. This limitation can lead to biased evaluations that do not fully represent an LLM's operational capacities, especially for larger LLMs requiring more intensive testing environments. In terms of integrity, the difficulty in installing and scaling benchmarks can create inconsistencies in how LLMs are evaluated, opening the door to manipulation. If a benchmark is too complex to implement consistently, results may vary between environments, allowing LLMs to be selectively tested in conditions favorable to them. This lack of standardization can compromise the benchmark's resistance to gaming by LLMs, undermining the reliability of its evaluation metrics. Improving the installation and scalability of benchmarks is crucial for ensuring that evaluations are both comprehensive and resistant to exploitation. Streamlined, well-documented benchmarks that can be easily deployed across different systems will enhance both the functional assessment of LLMs and the integrity of the evaluation process, enabling fairer and more accurate comparisons of model performance.

\subsection{Biases in LLM-Generated LLM Evaluations}
\label{subsec:biases-LLM-generated-evaluations}
\textbf{Prevalence}: 9/23

An emerging concern in LLM benchmarks is the use of LLMs themselves to generate evaluation tasks or assess other LLMs, effectively acting as both creator and judge \cite{gu2023xiezhi,li2023api,li2023halueval,liu2023agentbench,liu2023m3ke,sawada2023arb,srivastava2022beyond,tang2023toolalpaca,xu2023tool} (Appendix \ref{subsec:Appendix1-biases-LLM-generated-evaluations}). This practice introduces the risk of amplifying inherent biases and inaccuracies present in the evaluating LLMs, which can significantly undermine the benchmark's functionality and integrity. For instance, M3KE \cite{liu2023m3ke} utilized ChatGPT-generated multiple-choice questions to assess Chinese LLMs, inherently incorporating the biases of ChatGPT.

From a functionality standpoint, such biases can distort the evaluation process, leading to inaccurate assessments of LLM capabilities. For instance, an LLM-generated benchmark might inadvertently favor certain types of responses or reasoning patterns, resulting in an overestimation or underestimation of the evaluated model’s performance \cite{chang2023survey,min2023recent}. This issue raises concerns about the benchmark's ability to provide a true measure of an LLM’s diverse capabilities, potentially skewing its alignment with real-world applications. Regarding integrity, relying on LLMs for evaluation introduces the risk of circular reasoning, where the biases inherent in the LLMs generating the benchmarks could lead to evaluations that are not truly independent. This undermines the benchmark's resistance to manipulation, as LLMs could exploit these biases to achieve misleadingly high scores. For example, if an LLM-generated benchmark unintentionally reinforces specific patterns or knowledge it was trained on, it may allow LLMs to perform well through pattern recognition rather than demonstrating genuine understanding or reasoning \cite{ji2023survey,zhou2023don}. To ensure both functionality and integrity, a more balanced approach that integrates human expertise is necessary. Human involvement can mitigate the biases introduced by LLM-generated content, ensuring a more rigorous and objective evaluation process. This hybrid approach would enhance the benchmark's ability to provide a comprehensive and unbiased assessment of LLM capabilities, ensuring that evaluations remain reliable and reflective of true model performance.

\section{Processual Elements}
\label{sec:processual-elements}
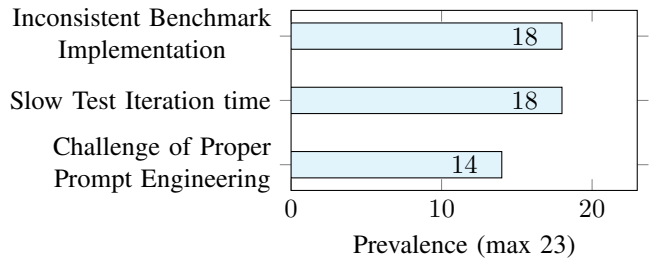
\begin{figure}[t!]
	\centering
	\resizebox{\columnwidth}{!}{
		\begin{tikzpicture}
			\begin{axis}[
				xbar, xmin=0, xmax=23,
				width=0.7\columnwidth,
				height=0.45\columnwidth, 
				enlarge y limits=0.2,
				xlabel={Prevalence (max 23)},
				symbolic y coords={
					{Challenge of Proper\\Prompt Engineering},
					{Slow Test Iteration time},
					{Inconsistent Benchmark\\Implementation}
				},
				ytick=data,
				yticklabel style={align=center},
				nodes near coords, nodes near coords align={horizontal},
				every node near coord/.append style={
					anchor=center,
					xshift=-0.5cm,
					color=black
				},
				]
				\addplot[fill=cyan!10]coordinates {
					(14,{Challenge of Proper\\Prompt Engineering})
					(18,{Slow Test Iteration time})
					(18,{Inconsistent Benchmark\\Implementation})
				};
			\end{axis}
		\end{tikzpicture}
	}
	\caption{Processual Inadequacies in LLM Benchmarking}
	\label{fig:Prevalence_Processual_Inadequacies}
\end{figure}

This section looks into the various process-related challenges and intricacies inherent in the implementation and evaluation of LLM benchmarks (Fig. \ref{fig:Prevalence_Processual_Inadequacies}). Please note that we use the word ``Processual'' to refer to the overall nature and development of processes, instead of ``procedural'' which focuses on specific steps or methods.

\subsection{Inconsistent Benchmark Implementation}
\label{subsec:inconsistent-benchmark-implementation-inconsistency}
\textbf{Prevalence}: 18/23

A critical issue identified in LLM benchmarks is the inconsistency in implementation across different research teams, as observed in \cite{chen2021evaluating,fu2023chain,gu2023xiezhi,guha2024legalbench,hendrycks2020measuring,li2023halueval,liang2022holistic,liu2023agentbench,liu2023m3ke,qin2023toolllm,sawada2023arb,shah2022flue,singhal2023large,tang2023toolalpaca,wang2023scibench,yu2023kola,zhong2024agieval,zhu2024promptbench} (Appendix \ref{subsec:Appendix2-benchmark-implementation-inconsistency}). Variations in the interpretation and execution of benchmarks, including differences in few-shot learning setups or chain-of-thought prompting, have led to diverse results that challenge the uniformity and reliability of these evaluations. For example, benchmarks like MMLU showed significant score variations due to differing implementation strategies \cite{ganguli2023challenges}. Similarly, M3KE \cite{liu2023m3ke} required substantial customization to accommodate language-specific subtleties, leading to inconsistencies in assessing Chinese LLMs across different research environments.

Functionally, this inconsistency hampers the benchmarks' ability to provide a uniform, objective assessment of LLMs. The lack of standardization in implementation can result in skewed outcomes, making it difficult to compare LLMs fairly or derive meaningful insights into their performance \cite{chang2023survey,ganguli2023challenges}. This undermines the benchmarks' functionality, as the results become less reliable for comparative analysis or performance tracking across different LLMs. From an integrity perspective, such inconsistencies can introduce vulnerabilities in the evaluation process. The absence of standardized implementation protocols opens up opportunities for manipulation or biased interpretations, where variations in methodology could be exploited to produce favorable outcomes without reflecting the model's genuine capabilities \cite{balloccu2024leak,zhou2023don}. This compromises the benchmark's integrity, as differing approaches might obscure true performance or create loopholes that can be exploited. To address such concerns, establishing standardized protocols and guidelines for benchmark implementation is essential. A uniform approach would enhance the consistency and reliability of LLM evaluations, ensuring that benchmarks provide a fair and accurate reflection of model capabilities.

\subsection{Slow Test Iteration Time}
\label{subsec:slow-iteration-time}
\textbf{Prevalence}: 18/23

Slow iteration time is a key inadequacy in LLM benchmarks, particularly in comprehensive third-party frameworks like BIG-bench and HELM \cite{fu2023chain,hendrycks2020measuring,huang2023c,li2023api,li2023halueval,liang2022holistic,liu2023agentbench,liu2023bolaa,sawada2023arb,shah2022flue,singhal2023large,srivastava2022beyond,tang2023toolalpaca,wang2023scibench,xu2023tool,yu2023kola,zhong2024agieval,zhu2024promptbench} (Appendix \ref{subsec:Appendix2-slow-iteration-time}). These frameworks often involve extensive evaluation processes, which can span weeks or months, hindering timely feedback on new LLMs. External involvement in evaluations adds further delays, requiring coordination and engineering support, as seen in benchmarks like BIG-bench \cite{srivastava2022beyond} and HELM \cite{liang2022holistic}. The need for comprehensive testing across diverse scenarios inherently extends evaluation time. 

Functionally, prolonged iteration times hinder the benchmarks' ability to provide current insights into LLM performance. For instance, SciBench \cite{wang2023scibench} involved manual data extraction and complex problem formulations, which significantly extended the evaluation process, making it difficult to keep pace with the rapid advancements in LLM development. With AI advancing rapidly, an LLM may undergo significant updates during its evaluation period, potentially rendering the results outdated or irrelevant by the time they are published \cite{chang2023survey,mcintosh2023google}. From an integrity perspective, slow iteration times pose security risks. Delayed evaluations mean emerging threats or vulnerabilities in LLMs may go undetected for extended periods, exposing systems to potential exploitation \cite{greshake2023not,wei2023jailbroken}. For example, ToolBench \cite{qin2023toolllm} integrated LLMs with a vast number of software APIs, requiring extensive manual configuration and adaptation, which could delay the identification and mitigation of security vulnerabilities. Additionally, the lag in benchmark feedback to LLMs would limit the ability to promptly address identified issues, allowing weaknesses to persist longer than necessary. To mitigate such challenges, streamlining evaluation processes and improving coordination with external parties is crucial. However, aligning the need for thorough evaluation with the rapid development pace of AI remains a complex issue, requiring innovative solutions to balance depth and timeliness in LLM benchmarking.

\subsection{Challenge of Proper Prompt Engineering in Benchmarking}
\label{subsec:prompt-engineering-challenge}
\textbf{Prevalence}: 14/23
Prompt engineering plays a crucial role in evaluating LLMs, as it shapes the interaction between the model and the test. While exploring different prompts is valuable for understanding LLM behavior, in the context of benchmarking, it is important to ensure that prompts are designed to accurately and consistently assess the LLM's capabilities. Challenges can arise when variations in prompt formulation lead to significant differences in model performance, which can complicate the interpretation of benchmark results \cite{fu2023chain,guha2024legalbench,hendrycks2020measuring,huang2023c,li2023halueval,liang2022holistic,liu2023agentbench,liu2023bolaa,liu2023m3ke,sawada2023arb,shah2022flue,singhal2023large,srivastava2022beyond,xu2023tool} (Appendix \ref{subsec:Appendix2-prompt-engineering-challenge}). For example, MMLU \cite{hendrycks2020measuring} employed a variety of prompts across different subjects to assess knowledge, but inconsistencies in prompt phrasing could lead to overestimation of the LLM's capabilities in certain areas. 

From a functionality perspective, inconsistency in prompt design can lead to assessments that do not accurately reflect the model's true capabilities. For example, prompts that are overly leading or ambiguous may result in responses that either overestimate or underestimate the LLM's performance, and this variability can make it difficult to compare results across different models or studies \cite{zamfirescu2023johnny}. Regarding integrity, if benchmarks do not standardize prompt construction, there is a risk that LLMs could be fine-tuned or engineered to perform well on specific prompt styles, potentially compromising the fairness of the evaluation \cite{deng2023multilingual,wei2023jailbroken}. To address such challenges, it is essential for benchmarks to adopt careful prompt engineering practices that aim for clarity, neutrality, and consistency, thereby ensuring that the evaluation accurately captures the model's capabilities without introducing unintended biases.

\section{Human Dynamics}
\label{sec:human-dynamics}
\begin{figure}[t!]
	\centering
	\resizebox{\columnwidth}{!}{
		\begin{tikzpicture}
			\begin{axis}[
				xbar, xmin=0, xmax=23,
				width=0.7\columnwidth,
				height=0.4\columnwidth, 
				enlarge y limits=0.25,
				xlabel={Prevalence (max 23)},
				symbolic y coords={
					{Diverse Cultural, Social,\\Political, Religious and\\Ideological Norms},
					{Diversity in Human\\Curators and Evaluators}
				},
				ytick=data,
				yticklabel style={align=center},
				nodes near coords, nodes near coords align={horizontal},
				every node near coord/.append style={
					anchor=center,
					xshift=-0.5cm,
					color=black
				},
				]
				\addplot[fill=cyan!10]coordinates {
					(18,{Diverse Cultural, Social,\\Political, Religious and\\Ideological Norms})
					(19,{Diversity in Human\\Curators and Evaluators})
				};
			\end{axis}
		\end{tikzpicture}
	}
	\caption{Human Dynamics Inadequacies in LLM Benchmarking}
	\label{fig:Prevalence_Human_Dynamics_Inadequacies}
\end{figure}
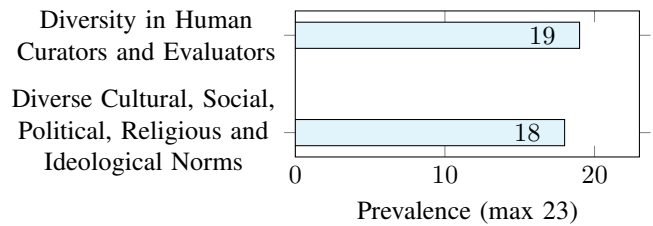

This section explores the subtle complexities of human factors influencing LLM benchmarks (Fig. \ref{fig:Prevalence_Human_Dynamics_Inadequacies}).

\subsection{Diversity in Human Curators and Evaluators}
\label{subsec:diversity-human-curators-evaluators}
\textbf{Prevalence}: 19/23

The heterogeneity among human curators and evaluators significantly impacts the construction and assessment of LLM benchmarks. Even with standardized guidelines, differences in cultural, religious, political, and academic or commercial backgrounds can introduce inconsistencies, particularly in subjective evaluations where linguistic and interpretative diversity play a role. This challenge was evident in benchmarks involving human judgment, such as those relying on sophisticated response evaluation or red-teaming (for cybersecurity purposes) \cite{chen2021evaluating,fu2023chain,gu2023xiezhi,guha2024legalbench,hendrycks2020measuring,huang2023c,li2023api,li2023halueval,liu2023agentbench,liu2023m3ke,qin2023toolllm,singhal2023large,srivastava2022beyond,tang2023toolalpaca,wang2023scibench,xu2023tool,yu2023kola,zhong2024agieval,zhu2024promptbench} (Appendix \ref{subsec:Appendix3-diversity-human-curators-evaluators}).

From a functionality perspective, this diversity can lead to subjective biases in benchmark construction and evaluation. For instance, cultural nuances can result in varying interpretations of responses, affecting the consistency and objectivity of the assessment \cite{chang2023survey,ganguli2023challenges,min2023recent}. For instance, LegalBench \cite{guha2024legalbench} utilized legal professionals from diverse jurisdictions to evaluate LLMs' legal reasoning, yet the varying interpretations of legal principles across different regions could skew the results and lead to inconsistent assessments, making it difficult to ensure a uniform evaluation standard across different LLMs. Regarding integrity, inconsistent application and interpretation by diverse evaluators can open avenues for manipulation or biased outcomes. Disparities in evaluation criteria might allow LLMs to exploit certain nuances to achieve favorable results, undermining the benchmark's robustness and trustworthiness \cite{greshake2023not}. This inconsistency can make it challenging to safeguard against potential biases or exploitation within the evaluation process. Addressing such issue requires a careful balance between leveraging diverse perspectives and ensuring consistency in evaluation. While diversity among evaluators enriches the assessment process, standardized protocols and training are necessary to minimize subjective biases and enhance the reliability of LLM benchmarks.

\subsection{Diverse Cultural, Social, Political, Religious, and Ideological Norms}
\label{subsec:diverse-cultural-ideological-challenges}
\textbf{Prevalence}: 18/23

LLM benchmarks face the inherent challenge of encompassing a wide range of cultural, religious, political, and ideological norms \cite{fu2023chain,gu2023xiezhi,guha2024legalbench,hendrycks2020measuring,huang2023c,li2023api,li2023halueval,liang2022holistic,liu2023agentbench,liu2023m3ke,sawada2023arb,shah2022flue,singhal2023large,srivastava2022beyond,wang2023scibench,yu2023kola,zhong2024agieval,zhu2024promptbench} (Appendix \ref{subsec:Appendix3-cultural-ideological-challenges}). Given that humans often disagree on fundamental issues, including interpretations of universal principles like human rights, pursuing a universal benchmark that addresses all cultural and ideological perspectives may be unrealistic. Benchmarks that rely on standardized answers or rubrics can inadvertently clash with diverse values, highlighting the impossibility of creating a one-size-fits-all evaluation. For example, LegalBench \cite{guha2024legalbench}, with its focus on American legal reasoning, may overestimate its applicability across diverse global legal contexts. Conversely, the integrity of Simplified Chinese LLM benchmarks like C-Eval \cite{huang2023c} can be easily gamed by specific prompt structures or leverage training data tailored to the Simplified Chinese context, such as offering responses that align with or praise Chinese ideologies, to achieve artificially high scores without demonstrating genuine understanding or reasoning \cite{yu2024cmoraleval}.

Functionally, this diversity means that benchmarks may struggle to fairly and accurately evaluate LLMs across varying viewpoints. An LLM's response to sensitive topics might be deemed appropriate in one cultural context but controversial in another \cite{mcintosh2023culturally,mcintosh2024inadequacy}. This inconsistency challenges the creation of benchmarks that can impartially assess LLMs' ability to navigate complex societal norms. An example of compromising benchmark integrity could be when benchmarks are tailored to avoid controversy (saying what people want to hear), leading to superficial assessments that do not truly evaluate the LLM’s ability to handle subtle or sensitive topics, thereby allowing LLMs to appear more competent than they actually are in real-world scenarios \cite{mcintosh2024inadequacy}. However, LLM providers and benchmark assessors are faced with the difficult decision of potentially offending some individuals, regardless of whether they aim for inclusivity or specificity. This is not a failure on their part, but rather an acknowledgment of the inherently subjective nature of humanity norms. Given such complexities, it may be more pragmatic for researchers to focus on context-specific benchmarks rather than striving for a universal standard benchmark for all human norms.

\section{Discussions}
\label{sec:discussions}
This section provides a comprehensive discussion of the underlying causes that have resulted in the previously aforementioned inadequacies in current LLM benchmarking methods.

\subsection{Misuse of the Term ``Benchmark''}
\label{subsec:misuse-benchmark-term}
The term \enquote{benchmark}, which should mean to the process of evaluating and comparing the performance of LLMs using standardized tests and metrics, has been applied more liberally in academic publishing than warranted, reflecting the evolving stage of AI regulatory compliance and public consensus on generative AI's emerging role and impact. Consequently, a wide array of non-standardized LLM test sets have been self-declared as LLM benchmarks, often lacking the rigor and uniformity required for true benchmarking. Genuine benchmarks should provide standardized, comprehensive evaluations, distinguishing them from more informal assessments rooted in researchers' subjective perspectives. Many so-called benchmarks serve merely as initial test sets tailored to specific LLM tasks, often falling short in integrity and functionality. Unlike fields such as automotive or aviation, where benchmarks are guided by established regulations and public consensus, the AI domain remains largely unregulated and in flux. This has led researchers to create varied task sets and questions, adding complexity to LLM evaluation and potentially misguiding interpretations of these sets as complete assessments of LLM capabilities. While such efforts contribute to the evolving landscape of LLM evaluation, highlighting the need for evolving standards, they also demonstrate the urgency for structured, universally accepted benchmarking frameworks. We believe that effective LLM benchmarks should rigorously assess functionality while being resilient to manipulation, ensuring they accurately reflect true model capabilities rather than superficial optimizations. The current research landscape requires benchmarks that align with emerging regulatory mandates and societal needs, particularly as LLMs become increasingly integrated into various aspects of life. Through rigorously designed benchmarks, we can ensure the accurate assessment of LLMs' growing capabilities, guiding their responsible development and deployment.

\subsection{Assessment Limitations for Reasoning and Multimodality}
\label{subsec:assessment-limitations}
Current LLM benchmarks exhibit significant limitations in both functionality and integrity when assessing comprehensive reasoning and multimodal capabilities. Functionally, many benchmarks fail to evaluate the depth of LLMs' reasoning and multimodal integration because they primarily focus on text-based tasks, neglecting the intricate processing required for integrating visual, auditory, and other sensory data \cite{chang2023survey,mcintosh2023google}. This narrow focus often results in benchmarks that assess only surface-level performance, without engaging the models in scenarios that require true multimodal understanding or complex reasoning. For example, benchmarks that rely on simple question-answer formats may inadvertently assess a model's ability to recall learned patterns rather than its capacity for creative problem-solving, thereby conflating crystal (recalled knowledge) and fluid (creative problem-solving) intelligence. In terms of integrity, such benchmarks are vulnerable to being manipulated by LLMs that have been overfitted to specific datasets or tuned to exploit particular evaluation metrics, when benchmarks allow models to achieve high scores through memorization or pattern exploitation rather than demonstrating genuine capability. For instance, a model trained on benchmark-specific data can deliver seemingly impressive results without possessing the underlying understanding or adaptability required in real-world applications. Consequently, current benchmarks may provide misleading assessments of LLM capabilities, due to superficial optimization strategies \cite{mcintosh2024inadequacy}.

\subsection{Unpredictability and Non-Repeatability}
\label{subsec:unpredictability}
The rapid evolution of commercial and open source LLMs introduces significant unpredictability and non-repeatability in benchmark assessments. From a functionality standpoint, the frequent updates and model variations released by vendors make it difficult to ascertain whether a benchmark measures an LLM's core capabilities or merely reflects transient optimizations tailored to the latest version. This instability undermines the benchmark's ability to consistently evaluate reasoning, comprehension, or multimodal integration, as the results may vary with each model iteration. Regarding integrity, the evolving nature of LLMs opens avenues for potential manipulation, as vendors can modify models to perform well on known benchmarks without genuinely improving underlying capabilities. This dynamic can lead to misleadingly high benchmark scores that do not represent the LLM's true functional performance in diverse, real-world scenarios. The lack of stable benchmarks complicates efforts to ensure that evaluations remain impartial and unaffected by superficial enhancements, thereby raising concerns about the authenticity of the assessments. Such issues require universally accepted, adaptable benchmarking frameworks that can accommodate the rapid advancements in LLM and generative AI, while providing reliable, repeatable measures of LLM performance. Without such standards, the evaluation landscape risks becoming fragmented and inconsistent, impeding the ability to accurately gauge the evolving capabilities of LLMs.

\subsection{Knowledge Boundaries in AI Benchmarking}
\label{subsec:knowledge-boundaries-ai}
LLM benchmarks are constrained by the current limits of the benchmark creators' human knowledge, hindering their ability to fully assess and cultivate emerging AI capabilities that may surpass conventional human understanding, potentially stalling innovation in fields that rely on advanced AI insights  \cite{mcintosh2024reasoning}. Additionally, the lack of specialized domain knowledge among benchmark evaluators can compromise the benchmark integrity, as generalist approaches often fail to address the subtle requirements of critical sectors such as national security or healthcare. Such deficiency not only introduces benchmarking biases, but also creates vulnerabilities, allowing LLMs to take advantage of ``benchmarks'' created by people who do not fully understand the specific field. To address such challenges, benchmarks must evolve to include a broader knowledge base and involve evaluators with deep expertise in specialized fields, which is crucial for creating assessments that genuinely reflect the advanced capabilities of LLMs and possibly soon Artificial General Intelligence (AGI), and are robust against manipulation. The rapid progression of LLMs requires LLM benchmarks that are as innovative and adaptable as the systems they evaluate, ensuring accurate measurement of LLM functionalities while maintaining integrity against evolving AI threats. Developing such benchmarks requires a concerted effort to integrate domain-specific knowledge into the evaluation process, ensuring that the assessments remain relevant and resilient in the face of AI's advancing frontier.

\subsection{Challenges in Inclusive Benchmarking}
\label{subsec:cultural-inclusivity-challenges}
Standardizing benchmarks for LLMs faces significant challenges due to the diversity of human values and perspectives, which extend beyond linguistic, cultural, religious, and political realms to encompass a broad spectrum of societal norms and ethical considerations, and are sometimes conflicting and irreconcilable. The predominance of English and Simplified Chinese (overlooking Traditional Chinese differences) in those benchmarks frequently reflected the default values and beliefs of their creators, overlooking the pluralistic nature of global beliefs and practices. For example, the English and Simplified Chinese centrism in benchmarks like BIG-Bench \cite{srivastava2022beyond}, and the focus on crystallized knowledge in MultiMedQA \cite{singhal2023large} and ARB \cite{sawada2023arb}, highlighted such challenges, potentially disadvantaging LLMs not trained in those more prevalent languages and limiting assessments of their broader reasoning capabilities. Such a narrower focus would compromise the benchmarks' ability to comprehensively and holistically evaluate LLMs, potentially favoring LLMs that aligned with specific cultural or ideological norms at the expense of broader applicability and acceptance, and raising concerns about the trustworthiness of LLMs in the absence of cultural and ideological inclusivity. The reliance on standardized answers or rubrics exacerbated this issue, especially in assessments designed to gauge fairness, societal appropriateness, or ethical decision-making, where the rich tapestry of human diversity demanded a more sophisticated approach. Addressing the aforementioned challenges would require moving beyond technical fixes to incorporate ethical decision-making and cultural sensitivity, highlighting the need for benchmarks that facilitate the development of globally aware LLMs, acknowledging the vast array of human values \cite{mcintosh2023culturally}. As the development of future LLM benchmarks progresses, it is anticipated that many will encounter similar challenges, particularly when the diversity of benchmark creators, evaluators, or the content itself does not fully embrace the broad spectrum of human diversity. However, different jurisdictions, religions, or cultures may have varying interpretations of inclusivity, ethics, or societal fairness that could lead to disagreements on what constitutes a superior LLM, which extends beyond technical benchmarks into broader considerations of values and politics.

\subsection{Limitations of This Study}
\label{subsec:limitations-study}
This study, while aiming to provide a thorough critique of LLM benchmarks, encounters its own set of limitations. Firstly, our analysis inherently carries our subjective bias, a challenge we have attempted to mitigate by transparently listing our evaluative rationale in the Appendices. Due to the varying level of detail in benchmark descriptions, when we could not conclusively determine whether a study had acknowledged or addressed a specific benchmark inadequacy, or evaluate the correctness of questions and answers (which themselves could become outdated as technology and modern civilization evolves, and the study in \cite{ganguli2023challenges} found mislabeled or unanswerable examples within MMLU \cite{hendrycks2020measuring}), we opted to give these studies the benefit of the doubt, acknowledging the potential for possible oversight. Secondly, while only 5 out of the 23 surveyed benchmark studies were peer-reviewed, our critical inclusion of preprints should not be seen as diminishing their scientific contribution, although it highlighted the evolving and dynamic nature of research in this field. Thirdly, our decision not to attempt reproducing results from the surveyed benchmark studies, was informed by the substantial time and effort required, coupled with the challenge of LLM updates potentially altering outcomes during the lengthy evaluation process. Furthermore, while we addressed linguistic differences between Simplified Chinese and Traditional Chinese, we did not explore the subtle distinctions among various English dialects (\textit{e.g.}, American, British) and their inherent logical differences related to cultural attitudes, thereby acknowledging another layer of complexity in language-based evaluations. Lastly, the absence of specific regulations and guidelines for generative AI and LLMs, and a lack of public consensus on acceptable AI behaviors, constrained our ability to definitively enumerate all benchmark inadequacies. The evolving capabilities of advanced AI systems would require ongoing amendments to our critique, reflecting the dynamic nature of the field and the continuous emergence of new AI features.

\subsection{Future Research Direction: Extending Benchmarks with Behavioral Profiling and Regular Audits}
\label{subsec:future-research-behavioral-profiling}
To effectively evaluate and utilize LLMs amid the swift advancements in generative AI, their evaluation methods can be extended beyond traditional benchmarks to include both initial screenings and in-depth, ongoing assessments that align with changing technological and societal needs. Similar to the initial candidate selection and aptitude tests in the human recruitment process, traditional benchmarks can serve as the first filter to screen out LLMs that fail to meet basic competence levels, akin to identifying candidates who might be fraudulent, incompetent, non-compliant or otherwise unsuitable. This step ensures that only LLMs with a foundational level of proficiency and regulatory compliance proceed to more rigorous evaluations, optimizing resource allocation for subsequent stages of the assessment process. 

Mirroring the interview stage of candidate selection, behavioral profiling explores deeper into the subtleties of LLM performance by assessing models beyond mere task completion, examining adaptability, ethical reasoning, and creativity in scenarios reflecting real-world complexities. Such AI behavioral analysis and profiling, a topic for future research, will prioritize LLMs most likely to fulfill specific roles or tasks, providing a sophisticated understanding of model capabilities and informing better selection decisions for deployment in sensitive or critical applications. Echoing the probation period in employment, regular audits post-deployment serve to continuously assess LLM performance against evolving standards and expectations. This step is crucial for catching any discrepancies between initial evaluations and actual performance, ensuring LLMs remain aligned with the requirements and ethical standards of evaluators and regulators over time. However, there are certain methodological limitations inherent in our approach. The subjective nature of benchmark evaluations, particularly when assessing more abstract qualities like adaptability or ethical reasoning, introduces potential bias. Additionally, the exclusion of certain benchmarks or evaluation frameworks may limit the generalizability of our findings. Future studies should address these gaps by exploring more objective, quantifiable methods for behavioral profiling and expanding the scope to include underrepresented benchmarks.

Given the rapid advancements in AI, our analysis risks becoming outdated as LLMs and benchmarking practices evolve. This highlights the need for ongoing research to continuously refine both the criteria and methods for LLM evaluation, ensuring they remain relevant and reflective of real-world complexities. Future research should focus on developing dynamic benchmarking systems that adapt to these changes, incorporating real-time audits and frequent updates to evaluation standards to capture the full scope of LLM capabilities and risks. Implementing this comprehensive evaluation framework will require collaboration across academia, industry, and regulatory bodies to develop standardized methodologies and ethical guidelines for each stage, ensuring assessments are rigorous and reflective of societal values and technological advancements. This will facilitate the creation of a dynamic evaluation ecosystem capable of adapting to the rapid pace of AI innovation and ensuring deployed LLMs are both effective and secure.

\section{Conclusion}
\label{sec:conclusion}
This study has critically analyzed 23 state-of-the-art LLM benchmarks, exposing significant inadequacies across technological, processual, and human dynamics that compromise their accuracy and reliability. The lack of standardized frameworks in AI benchmarking, unlike the established practices in regulated industries, has led to a proliferation of researcher-defined benchmarks that inadequately capture the complexity and evolving nature of LLMs. Our key contributions include: (i) formulating a unified evaluation framework rooted in cybersecurity principles to systematically identify and address deficiencies in benchmark design, focusing on functionality and integrity; (ii) conducting a detailed critique of 23 prominent LLM benchmarks, highlighting widespread issues that could impair comprehensive LLM assessment; and (iii) proposing the integration of LLM behavioral profiling and regular audits to enhance evaluation methodologies. As we arguably edge closer to the realization of AGI-like capabilities, it becomes crucial to rectify such benchmarking deficiencies, to ensure the responsible and secure deployment of LLMs. Furthermore, the lack of research consensus on how to properly benchmark LLMs, coupled with the academic liberty of releasing preprints, is likely to lead to a continued influx of question-and-answer sets self-claimed as benchmarks. It is therefore critical to point out their limitations and encourage deeper reflection within the research community. Our proposed extension of benchmarks with behavioral profiling and audits can provide a more subtle and rigorous evaluation of LLM capabilities and risks. Future work should prioritize the development of such benchmarks and establish standardized evaluation guidelines. We advocate for an international collaborative effort involving academia, industry, and regulatory bodies to continuously refine LLM benchmarks, aligning them with technological advancements and societal needs to ensure the development of robust and trustworthy AI systems.

\bibliographystyle{IEEEtran}
\bibliography{LLM_Benchmarking}

\vfill

%
%

\clearpage

\appendices

\section{Examples of Benchmark Inadequacies in Technological Aspects}
\label{sec:Appendix1}

\subsection{Response Variability in Standardized Evaluations}
\label{subsec:Appendix1-response-variability-standardized-evaluations}

\textbf{Prevalence}: 22/23

\begin{itemize}
	\item MMLU \cite{hendrycks2020measuring}: The study designed a benchmark that focused on a wide range of tasks across different fields, like Professional Law and College Medicine. However, it reported performance variations among different LLMs, particularly GPT-3, which struggled with tasks requiring detailed procedural knowledge or calculations, like Elementary Mathematics. This indicates the benchmark's potential inadequacy in capturing the specialized performance of LLMs in specific scenarios or formats.
	
	\item HumanEval \cite{chen2021evaluating}: The study demonstrated inadequacy in response variability as it evaluated Codex's performance mainly through functional correctness, using a unique set of hand-written programming problems (HumanEval), which might not fully capture the model's behavior in diverse real-world scenarios.
		
	\item LegalBench \cite{guha2024legalbench}: The study demonstrated inadequacies in standardized evaluations by applying generic benchmarks to specialized legal reasoning tasks, which did not accurately reflect the models' capabilities in the specific context of legal reasoning. This is exemplified in their development of the IRAC framework-based tasks, which sought to evaluate the application of legal reasoning but might not align with the specificities of the models tested.
	
	\item FLUE \cite{shah2022flue}: The study developed a novel Financial Language Model (FLANG) and Financial Language Understanding Evaluation (FLUE) benchmarks, but the benchmarks might not effectively represent model performance in specific financial contexts, as they focused on generic linguistic capabilities rather than specialized financial scenarios.
	
	\item MultiMedQA \cite{singhal2023large}: The study applied standardized benchmarks (MultiMedQA) to Flan-PaLM, a model not exclusively trained on medical data. This approach failed to fully capture Flan-PaLM's specialized performance in the medical context, as evidenced by its limited alignment with clinical consensus compared to clinician-generated answers.
	
	\item M3KE \cite{liu2023m3ke}: The study evaluated Chinese Large Language Models using the M3KE benchmark, which consists of multiple-choice questions across diverse subjects and education levels. However, this approach did not account for the specialized contexts and formats that specific models may be tailored for, thus not fully capturing the subtle performance of these models in their intended use cases.
	
	\item T-Bench \cite{xu2023tool}: The study evaluated LLMs on tool manipulation tasks with diverse software tools, incorporating both existing datasets and newly collected ones. It aimed to address the variability in model responses by enhancing open-source LLMs with techniques like model alignment with programmatic data curation, demonstration retrieval, and generation regulation with system prompts. However, the benchmark's design inherently might not fully capture the specialized, context-driven response behavior of models tailored for specific scenarios, reflecting the inadequacy related to response variability in standardized evaluations.
	
	\item Chain-of-Thought Hub \cite{fu2023chain}: The study introduced the Chain-of-Thought Hub for evaluating LLMs' reasoning capabilities, but \underline{acknowledged} the challenge of accurately representing models' performance due to variability in response behavior across different reasoning tasks and datasets.
	
	\item KoLA \cite{yu2023kola}: The study demonstrated this inadequacy by evaluating LLMs across a range of tasks designed to measure different types of knowledge (memorization, understanding, applying, creating), suggesting that the benchmark may not accurately capture model performance in specialized or intended contexts, particularly highlighted in its design of evolving and known data sources to assess models' handling of unseen data.
	
	\item SciBench \cite{wang2023scibench}: The study's benchmarks demonstrated the inadequacy of capturing response variability, when applied to models designed for specific contexts, as seen in their detailed evaluation of LLMs using varied prompting strategies and external tools, which failed to uniformly enhance model performance across different scientific problem-solving tasks.
	
	\item ARB \cite{sawada2023arb}: The ARB benchmark was designed to evaluate advanced reasoning capabilities in LLMs across various domains, including mathematics, physics, and law, with a focus on expert-level problem solving that requires specific formats or use cases. However, the study \underline{acknowledged} the challenge in accurately representing model performance in specialized contexts, particularly in symbolic reasoning and proof-like problems, which required sophisticated understanding and evaluation methodologies beyond standard metrics. This reflected the inadequacy in capturing the true performance and subtleties of models tailored for particular scenarios or formats.
	
	\item BIG-Bench \cite{srivastava2022beyond}: The study demonstrated the benchmark inadequacy of response variability when standardized evaluations were applied to models designed for specific formats, as evidenced by its extensive evaluation across a wide range of tasks and model scales without specifically tailoring benchmarks to individual model contexts or use cases.
	
	\item AGIEval \cite{zhong2024agieval}: The study designed AGIEval, a novel benchmark focused on human-centric tasks, but encountered challenges in accurately evaluating models across varied tasks, such as complex reasoning and domain-specific knowledge requirements, indicating a potential underestimation of model capabilities in specific contexts.
	
	\item ToolAlpaca \cite{tang2023toolalpaca}: The study introduced ToolAlpaca, a framework aiming to enhance compact language models' ability to generalize tool use by training on a diverse and simulated tool-use corpus. However, the evaluation primarily focused on the models' capability to handle unseen tools based on the generated dataset, without explicitly addressing or mitigating the variability in model responses to standardized evaluations for specific scenarios or formats, which could impact the accuracy of assessing a model's performance in its intended context.
	
	\item HELM \cite{liang2022holistic}: The study \underline{acknowledged} the challenge of standardizing language model evaluation across diverse scenarios, indicating an understanding of response variability but not fully addressing it through its methodology.
	
	\item ToolBench \cite{qin2023toolllm}: The study designed and utilized the ToolBench framework, which, although comprehensive, might not fully account for the variability in responses when applied to specialized, context-driven models, as evidenced by its broad approach encompassing 16,000+ real-world APIs without specific adjustments for models designed for niche applications.
	
	\item PromptBench \cite{zhu2024promptbench}: The study demonstrated that LLMs were not robust to adversarial prompts, highlighting a significant gap in evaluating models' performance in real-world, context-specific scenarios. This inadequacy was evident through the extensive evaluation of LLMs' vulnerability to various adversarial prompt attacks, which was not specifically addressed or resolved within the framework of the benchmark study itself.
	
	\item AgentBench \cite{liu2023agentbench}: The study introduced AgentBench to evaluate LLMs as agents across diverse real-world scenarios, but it inherently faced the challenge of response variability due to standardized evaluations not fully capturing the specialized, context-driven response behavior of LLMs tailored for specific scenarios, as evidenced by the significant performance disparity between commercial and open-source LLMs across different tasks.
	
	\item APIBank \cite{li2023api}: The study developed API-Bank, which, despite its comprehensive design to evaluate tool-augmented LLMs through a variety of APIs and dialogue scenarios, inherently faces challenges in accurately representing model performance across specialized, context-driven scenarios, as it aims to generalize across a thousand domains and multiple tool usage abilities, potentially overlooking the complex performance of LLMs in specific, real-world applications.
	
	\item C-Eval \cite{huang2023c}: The study introduced C-EVAL as a comprehensive Chinese evaluation suite, but \underline{acknowledged} variability in model responses, especially highlighting that specialized models might not be accurately assessed by generic benchmarks, as seen in their varied performance across different subjects and levels.
	
	\item BOLAA \cite{liu2023bolaa}: The study demonstrated that benchmarks like BOLAA, designed to evaluate LLM-augmented autonomous agents across various scenarios, did not fully capture the specialized response behavior of models tailored for specific tasks, leading to variability in model responses. This was evident in the experiments conducted across different LLMs and task complexities, where the performance of orchestrated multi-agent strategies (BOLAA) in the WebShop and HotPotQA environments highlighted the challenge of accurately assessing the sophisticated capabilities of specialized models.
	
	\item HaluEval \cite{li2023halueval}: The study designed a benchmark, HaluEval, which focused on evaluating LLMs' ability to recognize hallucinated content without tailoring to the specific contexts or formats of the evaluated models. This approach inherently overlooked the specialized response behaviors of models in particular scenarios, leading to potential inaccuracies in reflecting true model performance across diverse applications.
\end{itemize}

\subsection{Genuine Reasoning vs Technical Optimization}
\label{subsec:Appendix1-discerning-reasoning-optimization}
\textbf{Prevalence}: 22/23
\begin{itemize}
	\item MMLU \cite{hendrycks2020measuring}: The study highlighted the challenge of discerning genuine reasoning from technical optimization in LLMs, as evidenced by the varied performance across tasks and the reliance on pretraining, without a clear mechanism to ensure genuine understanding or reasoning capabilities.
	
	\item HumanEval \cite{chen2021evaluating}: The study relied on automatic unit tests to evaluate the functional correctness of code generated by LLMs, which did not directly address whether the model's solutions stemmed from genuine understanding, or were merely optimized for the test scenarios, thus not fully assessing genuine reasoning capabilities.
	
	\item LegalBench \cite{guha2024legalbench}: The study's focus on developing benchmarks for legal reasoning in LLMs highlighted challenges in ensuring that the models genuinely understood legal concepts, rather than merely optimizing to fit benchmark requirements, as evidenced by the varied performance across tasks and the utilization of chain-of-thought prompts to improve reasoning capabilities.
	
	\item FLUE \cite{shah2022flue}: The study introduced a novel pre-training methodology and evaluation benchmarks, without explicitly addressing the challenge of discerning genuine reasoning from technical optimization in LLMs, focusing primarily on performance improvements in financial NLP tasks.
	
	\item MultiMedQA \cite{singhal2023large}: The study \underline{acknowledged} the challenge in assessing whether LLM responses to medical questions are derived from genuine understanding or technical optimization, particularly highlighting the reliance on benchmark performances that might not fully capture the complex reasoning capabilities required in clinical contexts. This inadequacy remained unaddressed, as the study focused on improving accuracy through technical means such as instruction prompt tuning, without directly resolving the underlying issue of discerning genuine reasoning from optimization in LLM outputs.
	
	\item M3KE \cite{liu2023m3ke}: The study utilized a unified prompt approach for zero-shot and few-shot settings without explicitly addressing the differentiation between models' genuine comprehension and their ability to technically optimize for benchmark performance.
	
	\item T-Bench \cite{xu2023tool}: The study focused on boosting open-source LLMs' tool manipulation capabilities using methods like model alignment, demonstration retrieval, and generation regulation, without directly addressing the challenge of distinguishing genuine reasoning from technical optimization in LLM responses.
	
	\item Chain-of-Thought Hub \cite{fu2023chain}: The study used final answer accuracy as a proxy for reasoning capability without considering the correctness of intermediate steps, potentially failing to differentiate genuine reasoning from technical optimization.
	
	\item SciBench \cite{wang2023scibench}: The study demonstrated challenges in evaluating the true reasoning capabilities of LLMs due to reliance on technical optimization, as seen when LLMs, prompted with external tools, incorrectly derived equations, highlighting their struggle to understand mathematical relationships beyond mere optimization.
	
	\item ARB \cite{sawada2023arb}: The study utilized a rubric-based evaluation, which, despite its innovative approach, could not definitively resolve whether LLMs' responses stemmed from true understanding or were merely optimized for the test conditions.
	
	\item Xiezhi \cite{gu2023xiezhi}: The study developed a comprehensive, multi-disciplinary, auto-updating benchmark named Xiezhi, aimed at evaluating domain knowledge. However, it focused on creating a broad and balanced dataset without specifically addressing the challenge of discerning genuine reasoning from technical optimization in LLMs. The benchmark's design emphasizes the variety and freshness of questions but does not detail mechanisms for evaluating the depth of understanding versus surface-level pattern recognition.
	
	\item BIG-Bench \cite{srivastava2022beyond}: The study introduced BIG-Bench, which includes tasks beyond current LLM capabilities, aiming to better understand and predict LLM behavior. However, it inherently struggles to differentiate between genuine reasoning and technical optimization due to the diverse and complex nature of tasks, some of which may allow models to optimize for benchmark performance without demonstrating true comprehension.
	
	\item AGIEval \cite{zhong2024agieval}: The study utilized the Chain-of-Thought prompting technique to assess models' reasoning capabilities, but it noted variability in performance improvements across tasks, suggesting a challenge in conclusively determining genuine reasoning abilities versus technical optimization.
	
	\item ToolAlpaca \cite{tang2023toolalpaca}: The study created and evaluated models using simulated tool-use scenarios, without explicitly addressing the distinction between genuine reasoning and optimization to benchmark specifications, exemplified by the training and evaluation of models on a constructed corpus without mechanisms to differentiate between true comprehension and mere performance optimization.
	
	\item HELM \cite{liang2022holistic}: The HELM evaluation framework explicitly aimed to address various facets of language model capabilities, including reasoning, but the inherent challenge of distinguishing between genuine reasoning and technical optimization in responses remains a significant concern due to the models' \enquote{black box} nature and the complexity of interpreting LLM outputs beyond mere performance metrics.
	
	\item ToolBench \cite{qin2023toolllm}: The study implemented a depth-first search-based decision tree (DFSDT) to enhance the planning and reasoning ability of LLMs, indicating an effort to discern genuine reasoning from technical optimization. However, the primary focus was on improving LLMs' ability to interact with APIs rather than directly addressing the benchmark's potential inadequacy in distinguishing genuine reasoning capabilities from mere technical optimization.
	
	\item PromptBench \cite{zhu2024promptbench}: The study \underline{acknowledged} the challenge of discerning genuine reasoning from technical optimization in LLMs, indicating that it primarily evaluated models based on their ability to follow instructions and generate outputs without deeply investigating whether these outputs resulted from true understanding or merely optimization strategies. This suggested that the benchmarks could not fully distinguish between genuine reasoning and technical optimization, reflecting the described inadequacy.
	
	\item AgentBench \cite{liu2023agentbench}: The study did not provide explicit methods to differentiate whether LLMs' responses were the result of genuine reasoning or merely technical optimizations, focusing instead on evaluating LLMs' ability to act as agents in various environments without addressing this specific concern.
	
	\item APIBank \cite{li2023api}: The study detailed an evaluation system for tool-augmented LLMs, without directly addressing the challenge of determining whether LLM responses stem from genuine reasoning or merely technical optimization to match benchmark answers.
	
	\item C-Eval \cite{huang2023c}: The study emphasized evaluating advanced reasoning abilities through C-EVAL HARD with complex questions, but it did not specifically address or provide mechanisms to distinguish between genuine understanding and technical optimization by LLMs, focusing instead on the broad assessment of reasoning abilities without delving into the nature of the reasoning process itself.
	
	\item BOLAA \cite{liu2023bolaa}: The study did not specifically address the challenge of discerning genuine reasoning from technical optimization in LLMs, focusing instead on evaluating LAAs across different architectures and tasks, without explicitly examining the underlying reasoning capabilities of the LLMs involved.
	
	\item HaluEval \cite{li2023halueval}: The study's methodology, focused on evaluating LLMs' ability to recognize hallucinated content, indirectly involves assessing models based on their output, without thoroughly investigating the underlying reasoning processes, or the extent to which responses are generated through comprehension versus optimization for benchmark performance.
\end{itemize}

\subsection{Tension Between Helpfulness and Harmlessness}
\label{subsec:Appendix1-tension-helpfulness-harmlessness}
\textbf{Prevalence}: 19/23
\begin{itemize}
	\item MMLU \cite{hendrycks2020measuring}: The study \underline{acknowledged} the challenge in balancing helpfulness with harmlessness across various tasks, including those with significant societal relevance like law and morality, without offering a resolved methodology to accurately gauge this balance within its benchmarks.
	
	\item HumanEval \cite{chen2021evaluating}: The study focused on evaluating LLMs trained on code, specifically assessing their ability to generate standalone Python functions from docstrings. However, it did not address the tension between helpfulness and harmlessness in LLM benchmarks. The methodology primarily centered on functional correctness through unit tests and did not incorporate a framework to evaluate or balance the helpfulness versus harmlessness of the generated code, which is essential in diverse real-world scenarios.
	
	\item LegalBench \cite{guha2024legalbench}: The study did not provide a standardized criterion for balancing helpfulness and harmlessness in evaluating LLMs, focusing instead on legal reasoning tasks without addressing this specific tension.

	\item FLUE \cite{shah2022flue}: The study utilized A/B tests for human evaluation without explicitly addressing the balance between helpfulness and harmlessness, potentially leading to an overemphasis on one aspect over the other in model evaluation.
	
	\item MultiMedQA \cite{singhal2023large}: The study introduced a sophisticated approach for evaluating LLMs in medical question answering, incorporating human evaluation frameworks to assess the balance between helpfulness and harmlessness; however, it recognized the complexity in achieving this balance due to the intricate nature of medical knowledge and societal values, without presenting a resolved methodology to adequately quantify or standardize this balance.
	
	\item M3KE \cite{liu2023m3ke}: The study did not establish a clear criterion for balancing helpfulness and harmlessness in evaluating LLMs, focusing instead on assessing knowledge across various subjects and education levels without addressing the subtle balance between generating useful and non-harmful responses.
	
	\item T-Bench \cite{xu2023tool}: The study improved open-source LLMs' tool manipulation capabilities using techniques like model alignment, system prompts, and demonstration retrievers, without specifically addressing the balance between helpfulness and harmlessness.
	
	\item Chain-of-Thought Hub \cite{fu2023chain}: The study did not establish a clear criterion for balancing helpfulness and harmlessness in its evaluation of LLMs, focusing instead on reasoning capabilities without addressing the potential for harmful outputs in diverse real-world scenarios.
	
	\item KoLA \cite{yu2023kola}: The study introduced a contrastive evaluation system focusing on knowledge hallucination and overall performance, without explicitly addressing the balance between helpfulness and harmlessness in its methodology.
	
	\item SciBench \cite{wang2023scibench}: The study demonstrated a tension between helpfulness and harmlessness in LLM benchmarks by emphasizing benchmarks' struggle to quantify a balance between providing helpful information and avoiding harmful outputs, particularly through the use of A/B tests and the incorporation of human evaluation methods, without a clear criterion for this balance.
	
	\item ARB \cite{sawada2023arb}: The study did not specifically address or demonstrate unresolved inadequacies regarding the tension between helpfulness and harmlessness in LLM benchmarks. It focused on evaluating LLMs' advanced reasoning capabilities across various domains, without explicitly discussing the balance between providing useful information and avoiding harmful outputs in its benchmarking approach.
	
	\item AGIEval \cite{zhong2024agieval}: The study utilized a human-centric benchmark based on standardized exams, without explicitly addressing the balance between generating helpful and non-harmful outputs in diverse real-world contexts.
	
	\item HELM \cite{liang2022holistic}: The HELM evaluation framework included targeted evaluations and multi-metric measurement across various scenarios, but did not specifically address the tension between helpfulness and harmlessness in its criteria or methodologies, indicating an unresolved challenge in balancing these aspects within the 
	
	\item ToolBench \cite{qin2023toolllm}: The study introduced ToolBench and DFSDT to enhance LLMs' tool-use capabilities, but did not specifically address the balance between helpfulness and harmlessness, focusing instead on tool integration and reasoning efficiency without detailing measures to avoid harmful outputs.
	
	\item PromptBench \cite{zhu2024promptbench}: The study did not establish a clear criterion for balancing helpfulness and harmlessness in adversarial prompt evaluation, focusing instead on the robustness of LLMs against adversarial prompts without directly addressing how these models balance being helpful and avoiding harmful outputs in real-world scenarios.
	
	\item AgentBench \cite{liu2023agentbench}: The study extensively evaluated LLMs across various tasks and environments, but did not establish a clear methodology or criteria for balancing the provision of useful information with the prevention of harmful outputs, as seen in their evaluation setup and performance analysis sections. This gap indicated a potential oversight in considering the complex and subtle balance between those aspects in real-world applications, directly impacting the benchmark's ability to comprehensively evaluate LLMs for both functionality and cybersecurity.
	
	\item C-Eval \cite{huang2023c}: The study implemented a comprehensive evaluation across diverse disciplines, but it did not specifically address the balance between helpfulness and harmlessness, particularly in the context of cybersecurity threats, indicating a gap in ensuring LLMs' responses are both useful and non-harmful.
	
	\item BOLAA \cite{liu2023bolaa}: The experiment focused on orchestrating multiple LAAs to enhance decision-making and knowledge reasoning capabilities, without specifically addressing how these models manage the trade-off between generating helpful responses and avoiding harmful outputs in diverse real-world scenarios.
	
	\item HaluEval \cite{li2023halueval}: The study introduced HaluEval, a benchmark for evaluating LLMs' performance in recognizing hallucinations, which inherently did not address the tension between helpfulness and harmlessness. It focused on generating and annotating hallucinated responses to assess LLMs' ability to recognize hallucinations, without directly tackling the balance between providing useful information and avoiding harmful outputs in the benchmark design.
\end{itemize}

\subsection{Linguistic Variability and Embedded Logic Diversity}
\label{subsec:Appendix1-linguistic-differences-embedded-logics}
\textbf{Prevalence}: 17/23

\begin{itemize}
	\item MMLU \cite{hendrycks2020measuring}: The study introduced a benchmark covering a wide range of subjects but did not specifically address or incorporate the diversity of cognitive and logical frameworks across different languages, focusing instead on assessing language understanding primarily through English-based tasks. The study \underline{acknowledged} the challenge of evaluating LLMs across diverse knowledge domains, but did not explicitly attempt to address the linguistic diversity inadequacy.
	
	\item HumanEval \cite{chen2021evaluating}: The study focused exclusively on code generation from English docstrings, without considering or addressing the subtleties of linguistic diversity or the embedded logic differences in various languages. This approach inherently ignored the benchmark inadequacy related to linguistic differences and embedded logics, as it evaluated the LLM's code generation capability solely based on English, without any mention of adaptation or evaluation across different languages or addressing the inherent cognitive frameworks shaped by each language's unique structure. The benchmark study did not acknowledge this inadequacy or make attempts to address it.
	
	\item LegalBench \cite{guha2024legalbench}: The study focused on developing and evaluating benchmarks within the context of American legal reasoning and did not address linguistic differences or embedded logics in different languages. It utilized tasks primarily in \enquote{Natural English}, without considering multilingual contexts or the diverse cognitive frameworks shaped by each language's unique structure. The benchmark study did not acknowledge this specific inadequacy or make attempts to address it.
	
	\item FLUE \cite{shah2022flue}: The study focused on developing and evaluating language models specifically for the financial domain, utilizing primarily English datasets without addressing or incorporating multilingual contexts or the diverse cognitive frameworks shaped by different languages' unique structures. There was no mention of efforts to include or evaluate the models' performance across various languages or to consider linguistic diversity explicitly.
	
	\item MultiMedQA \cite{singhal2023large}: The study's benchmarks primarily focused on English-language datasets and did not address the need for multilingual evaluations, indicating a disregard for linguistic differences and embedded logics in different languages. While the benchmark study, MultiMedQA, was praised for its diversity and inclusion of various medical exam, research, and consumer sources, the study \underline{acknowledged} its limitation by not incorporating a broader variety of languages and cultural contexts, thereby not addressing the mentioned benchmark inadequacy.
	
	\item M3KE \cite{liu2023m3ke}: The benchmark study focused exclusively on Simplified Chinese LLMs, erroneously labeling them as \enquote{Chinese} without differentiation with Traditional Chinese, and did not address or recognize the diversity of cognitive and logical frameworks across languages, thus overlooking linguistic differences and embedded logics.
	
	\item T-Bench \cite{xu2023tool}: The study focused on enhancing open-source LLMs for tool manipulation without addressing the incorporation or evaluation of multilingual capabilities or considering linguistic diversity in its benchmarking process. The benchmark study, ToolBench, did not acknowledge or attempt to address this particular inadequacy of LLM benchmarks.
	
	\item Chain-of-Thought Hub \cite{fu2023chain}: The study included bilingual reasoning capabilities in English and Simplified Chinese (erroneously labeling them as \enquote{Chinese} without differentiation with Traditional Chinese), but did not specifically address or consider the linguistic differences and embedded logics across different languages beyond this, indicating a potential oversight of the depth of cognitive and logical frameworks unique to each language. The benchmark study did not acknowledge this inadequacy or make attempts to address it.
	
	\item KoLA \cite{yu2023kola}: The study \underline{acknowledged} the inadequacy by incorporating evolving data sources, including non-English content like news articles and novels, to evaluate models' understanding and generation abilities in diverse linguistic contexts. However, it did not explicitly address or attempt to resolve the benchmark inadequacy of ignoring linguistic differences and embedded logics in different languages.

	\item Xiezhi \cite{gu2023xiezhi}: The study constructed the Xiezhi benchmark focusing on domain knowledge evaluation across a wide array of disciplines without explicitly addressing linguistic diversity or the embedded logics in different languages. It predominantly utilized data from Simplified Chinese (erroneously labeling them as \enquote{Chinese} without differentiation with Traditional Chinese) educational systems and academic surveys, with no clear mention of adjustments or considerations for linguistic differences beyond Chinese and English, nor did it acknowledge this as a limitation or attempt to address it.
	
	\item AGIEval \cite{zhong2024agieval}: The study utilized a bilingual (English and Simplified Chinese, erroneously labeling them as \enquote{Chinese}) approach without specifically addressing or accounting for the unique cognitive and logical frameworks shaped by each language's structure, thereby not fully addressing the benchmark inadequacy regarding linguistic diversity and embedded logic. While the study \underline{acknowledged} the inclusion of tasks in both English and Chinese, it did not discuss efforts to address the unique systems of reasoning fostered by different linguistic structures, suggesting a potential overlook of the linguistic diversity and embedded logic inadequacy.
	
	\item ToolAlpaca \cite{tang2023toolalpaca}: The study predominantly utilized English for instructions and API documentation generation without explicitly addressing or incorporating linguistic differences or the unique logics embedded in various languages. The study did not acknowledge or attempt to address the benchmark inadequacy related to ignoring linguistic differences and embedded logics in different languages.
	
	\item HELM \cite{liang2022holistic}: The study \underline{acknowledged} the challenge of accounting for linguistic diversity in benchmarks, but did not provide a solution, exemplified by its focus on English and limited consideration of other languages, which indicates an unaddressed and unresolved inadequacy regarding linguistic differences and embedded logics.
	
	\item ToolBench \cite{qin2023toolllm}: The study focused on integrating LLMs with a vast number of real-world APIs without explicitly addressing the challenge of linguistic diversity or the embedded logics in different languages. The methodology was primarily based on automatic construction using ChatGPT, which inherently follows a predominantly English-centric approach due to its training data and design, without adapting to or considering the unique cognitive and logical frameworks that different languages might embody. The study did not acknowledge this specific inadequacy or attempt to address it.
	
	\item APIBank \cite{li2023api}: The study's benchmark, API-Bank, was primarily focused on English and did not address the construction and evaluation of models for other languages, indicating a disregard for linguistic differences and embedded logics in different languages. The benchmark study \underline{acknowledged} this limitation and mentioned plans to address data construction and model evaluation for other languages as future work.
	
	\item C-Eval \cite{huang2023c}: The study explicitly focused on evaluating LLMs in a Simplified Chinese (erroneously labeling them as \enquote{Chinese} without differentiation with Traditional Chinese) context, without addressing or attempting to include diverse linguistic backgrounds or cognitive frameworks, thus overlooking the benchmark inadequacy related to ignoring linguistic differences and embedded logics in different languages. The benchmark did not acknowledge or address this inadequacy.
	
	\item HaluEval \cite{li2023halueval}: The study focused on evaluating LLMs' ability to recognize hallucinations across different tasks, without addressing the linguistic diversity or embedded logic specific to non-English languages. The benchmark did not acknowledge or attempt to address this inadequacy, relying instead on predominantly English-centric data and evaluation criteria.
\end{itemize}

\subsection{Benchmark Installation and Scalability}
\label{subsec:Appendix1-benchmark-installation-scalability}

\textbf{Prevalence}: 16/23

\begin{itemize}
	\item MMLU \cite{hendrycks2020measuring}: The study demonstrated challenges in scalability and installation complexity for benchmarks, as evidenced by the extensive engineering efforts and resource allocation required to implement and scale the benchmark across various scenarios and models.
	
	\item LegalBench \cite{guha2024legalbench}: The study encountered challenges in efficiently scaling its benchmark across various scenarios and models, as evidenced by the need for manually generating and annotating samples for specific tasks like the \enquote{Hearsay} or \enquote{Personal Jurisdiction} sections.
	
	\item M3KE \cite{liu2023m3ke}: The study introduced M3KE, requiring significant engineering efforts for setting up and assessing a wide range of Chinese LLMs, without specifically addressing the complexity and scalability challenges of the benchmark's installation and operational efficiency.
	
	\item T-Bench \cite{xu2023tool}: The study encountered substantial difficulty in effectively scaling benchmarks for open-source LLMs. For instance, they noted significant performance disparities between open-source models and proprietary ones like GPT-4 in tool manipulation tasks, necessitating enhanced techniques and infrastructure to bridge this gap.
	
	\item KoLA \cite{yu2023kola}: The study required significant effort to install and scale the KoLA benchmark, as indicated by the detailed description of the evolving and known data sources and complex evaluation criteria, demonstrating the technological challenges inherent in this process.
	
	\item SciBench \cite{wang2023scibench}: The study demonstrated complexity in implementing and scaling benchmarks, as it involved intricate processes like manual data extraction from PDFs to LaTeX, complex problem formulations, and the use of external tools for evaluation, which aligned with the stated concerns of installation and scalability challenges.
	
	\item ARB \cite{sawada2023arb}: The study faced significant difficulties in efficiently scaling and implementing the ARB benchmark, evident in the extensive manual grading required for complex symbolic and proof-like problems, reflecting challenges in installation and scalability.
	
	\item Xiezhi \cite{gu2023xiezhi}: The study utilized the Xiezhi benchmark, which required considerable effort to adapt and scale for evaluating various LLMs, as evidenced by the detailed description of its complex setup and evaluation processes.
	
	\item BIG-Bench \cite{srivastava2022beyond}: The study encountered challenges in efficiently scaling the BIG-bench framework across various models, as evidenced by their focus on analyzing performance differences across model sizes and architectures.
	
	\item AGIEval \cite{zhong2024agieval}: The study encountered challenges in installing and scaling benchmarks for evaluating LLMs, as evidenced by the extensive use of different models (GPT-4, ChatGPT, Text-Davinci-003) and varied testing methodologies (zero-shot, few-shot, Chain-of-Thought), indicating significant engineering efforts for implementation.
	
	\item ToolAlpaca \cite{tang2023toolalpaca}: The study faced challenges in installing and scaling the framework for evaluating language models, as evidenced by their reliance on complex multi-agent simulation environments and sophisticated infrastructure for ToolAlpaca's implementation.
	
	\item PromptBench \cite{zhu2024promptbench}: The study encountered difficulties in efficiently scaling their benchmark framework, PromptBench, across various large language models, indicating substantial technical challenges in installation and scalability.
	
	\item AgentBench \cite{liu2023agentbench}: The study demonstrated the complexity of setting up and operating the AgentBench benchmark, which required extensive configuration and adaptation to various models and environments, indicating the challenges in installation and usage.
	
	\item APIBank \cite{li2023api}: The study faced significant challenges in installing and scaling the benchmark framework, evidenced by the complex implementation of 73 APIs, requiring substantial engineering efforts and extensive annotation processes.
	
	\item BOLAA \cite{liu2023bolaa}: The study encountered challenges in scaling and installing the benchmark framework for evaluating LLM-augmented Autonomous Agents (LAAs), necessitating considerable adjustments in infrastructure and resource allocation.
	
	\item HaluEval \cite{li2023halueval}: The study faced significant challenges in implementing and scaling their HaluEval, which required extensive use of resources and intricate engineering efforts, exemplified by their complex two-step generation process for hallucinated samples.	
\end{itemize}

\subsection{Biases in LLM-Generated LLM Evaluations}
\label{subsec:Appendix1-biases-LLM-generated-evaluations}

\textbf{Prevalence}: 9/23

\begin{itemize}
	
	\item M3KE \cite{liu2023m3ke}: The M3KE benchmark utilized model-generated multiple-choice questions to assess the capabilities of Chinese Large Language Models, inherently incorporating the biases and inaccuracies of the models used to generate these evaluations, as evidenced by their methodology of collecting and organizing questions from public websites and ensuring a standardized assessment process without explicitly addressing the mitigation of such biases in the creation or selection of these questions.
	
	\item T-Bench \cite{xu2023tool}: The study utilized generative AI models to enhance open-source LLMs for tool manipulation, explicitly involving model-generated evaluations through programmatic data generation and in-context demonstration retrievers, which could inherit biases from the models used to generate them.
	
	\item ARB \cite{sawada2023arb}: The study utilized GPT-4 to generate rubrics and evaluate the reasoning of symbolic math and proof-like problems, inheriting potential biases of GPT-4 in the evaluation process.
	
	\item Xiezhi \cite{gu2023xiezhi}: The benchmark utilized ChatGPT for tagging questions with disciplinary labels, demonstrating an instance where biases inherent to the ChatGPT model generating evaluations could have influenced the objectivity and reliability of the evaluations.
	
	\item BIG-Bench \cite{srivastava2022beyond}: The benchmark utilized generative AI models to assess other LLMs, inheriting the biases and inaccuracies from the models used to generate evaluations, thus potentially compromising the objectivity and reliability of these assessments.
	
	\item ToolAlpaca \cite{tang2023toolalpaca}: The study utilized generative AI models to automatically generate structured documentation for APIs, which were then used to simulate tool-use instances for training LLMs, inherently incorporating biases from the generative models used.
	
	\item AgentBench \cite{liu2023agentbench}: The study utilized generative AI models to create novel tasks and evaluate LLMs, inheriting biases from the models used for generation, as evidenced by the use of model-generated tasks in various environments without addressing the potential for inherited biases.
	
	\item APIBank \cite{li2023api}: The study utilized generative AI models to automatically mass-produce training data, which likely inherited biases from the models used, compromising the objectivity and reliability of evaluations.
	
	\item HaluEval \cite{li2023halueval}: The study utilized generative AI models, specifically ChatGPT, to automatically generate hallucinated samples for evaluation, inherently subjecting the benchmark to biases and inaccuracies of the generative model used, as described in the study's methodology for generating and evaluating hallucinated content.
\end{itemize}

\newpage

\section{Examples of Benchmark Inadequacies in Processual Elements}
\label{sec:Appendix2}
\setcounter{subsection}{0}

\subsection{Inconsistent Benchmark Implementation}
\label{subsec:Appendix2-benchmark-implementation-inconsistency}
\textbf{Prevalence}: 18/23

\begin{itemize}
	\item MMLU \cite{hendrycks2020measuring}: The study developed a new benchmark to evaluate LLMs across a diverse set of subjects and tested them in zero-shot and few-shot settings, indicating an innovative approach yet not directly addressing the uniformity and objectivity in benchmark implementation across different laboratories. 
	
	\item HumanEval \cite{chen2021evaluating}: The study \underline{acknowledged} the potential for diverse outcomes from varying implementation methods, but did not specifically address or propose solutions to standardize benchmark execution across different research teams, focusing instead on the development and evaluation of models using their own benchmarks and methodologies.
	
	\item LegalBench \cite{guha2024legalbench}: The study introduced LegalBench with an emphasis on diverse legal reasoning tasks and conducted preliminary evaluations on different models, yet it did not address the inconsistency in benchmark implementation directly. Although the study aimed to foster standardized evaluations through its collaborative and open design, it implicitly \underline{acknowledged} the challenge of ensuring uniformity across implementations by inviting community contributions to refine and expand the benchmark suite.
	
	\item FLUE \cite{shah2022flue}: The study introduced the Financial Language Understanding Evaluation (FLUE) and a new financial language model (FLANG), focusing on domain-specific pre-training and evaluation without explicitly addressing or attempting to standardize the implementation process across different teams or laboratories, which could lead to inconsistent benchmark applications and results. The study did not mention any measures to address this benchmark inadequacy directly.
	
	\item MultiMedQA \cite{singhal2023large}: The study introduced MultiMedQA, aiming to overcome limitations of existing benchmarks by combining six medical question answering datasets and a new dataset, HealthSearchQA, for comprehensive LLM evaluation. However, it recognized the challenge of ensuring consistent benchmark application across diverse medical domains, \underline{acknowledging} the difficulty in uniform implementation without directly addressing a solution to this inconsistency.
	
	\item M3KE \cite{liu2023m3ke}: The study developed M3KE, a benchmark for Simplified Chinese (erroneously labeling them as \enquote{Chinese} without differentiation with Traditional Chinese) LLMs, without addressing the inadequacy of inconsistent benchmark implementation; it did not mention any standardized protocols or guidelines to ensure uniformity across different research teams or laboratories.
	
	\item Chain-of-Thought Hub \cite{fu2023chain}: The study integrated diverse benchmarks (\textit{e.g.}, GSM8k, MATH, MMLU) for evaluating LLM reasoning, but did not address implementation inconsistencies, focusing instead on model performance comparison. There was no mention of efforts to standardize benchmark execution or address potential inconsistencies across different research teams.
	
	\item KoLA \cite{yu2023kola}: The study established a new benchmark, KoLA, designed to evaluate LLMs across various knowledge-intensive tasks using both known and evolving data sources, and introduced a contrastive evaluation system aimed at ensuring fairness and applicability in LLM assessment. However, it did not specifically address the inconsistency in benchmark execution across different laboratories, focusing instead on creating a more reliable and fair evaluation framework through innovative data sources and evaluation metrics. The benchmark study \underline{acknowledged} the need for better fairness and applicability in LLM evaluations but did not make explicit attempts to address the inconsistency in execution across different research teams.
	
	\item SciBench \cite{wang2023scibench}: The study introduced SciBench to systematically examine complex scientific problem-solving capabilities of LLMs, but it encountered inconsistencies in benchmark implementation. The benchmarks, derived from collegiate-level textbooks and exams, required advanced computations like integration and differentiation, which varied in execution complexity. Despite efforts to minimize data leakage and use detailed solutions for error analysis, the study \underline{acknowledged} the challenge of ensuring uniform assessment across different LLM configurations and the potential for varied interpretations and implementations of benchmarks, without detailing explicit measures to address these inconsistencies directly.
	
	\item ARB \cite{sawada2023arb}: The study introduced a novel benchmark, ARB, to evaluate LLMs using advanced reasoning problems, yet \underline{acknowledged} the challenge of ensuring consistent and reliable evaluation methods, particularly for symbolic and proof-like problems, without directly addressing the inconsistency in benchmark implementation across different labs.
	
	\item Xiezhi \cite{gu2023xiezhi}: The study introduced the Xiezhi benchmark, which aimed to establish a comprehensive and multi-disciplinary auto-updating benchmark. Despite its efforts to address inconsistencies by incorporating a broad range of disciplines and updating content, the study does not explicitly discuss measures taken to standardize benchmark implementation across different research teams or laboratories to ensure uniform execution and assessment.
	
	\item AGIEval \cite{zhong2024agieval}: The study employed a novel benchmark, AGIEval, which was specifically designed to evaluate foundation models on human-centric tasks derived from high-standard admission and qualification exams. The benchmarks aimed to assess models' general abilities in tasks related to human cognition and problem-solving. However, the study did not address the issue of implementation inconsistency across different laboratories. While the study made a significant effort to create a standardized and objective assessment through AGIEval, it did not explicitly mention measures to ensure uniform implementation especially of human-centric tasks across different research teams, which could potentially lead to variability in results similar to those observed in benchmarks like MMLU and BBQ.
	
	\item ToolAlpaca \cite{tang2023toolalpaca}: The study developed a novel framework, ToolAlpaca, to automatically generate a diverse tool-use corpus and enhance compact language models' generalized tool-use abilities, without specifically addressing or resolving the inconsistency in benchmark implementation across different laboratories. The benchmark study did not acknowledge this inadequacy of LLM benchmarks nor made attempts to address it.
	
	\item HELM \cite{liang2022holistic}: The study \underline{acknowledged} the challenge of ensuring uniform evaluation methods across all models, highlighting ongoing efforts to refine and adapt benchmarking practices to maintain relevancy and accuracy in assessments, without explicitly stating that this inadequacy has been fully resolved.
	
	\item ToolBench \cite{qin2023toolllm}: While the study introduced innovative methods like DFSDT to improve LLM's planning and reasoning abilities and developed an automatic evaluator, ToolEval, to assess LLM's tool-use capabilities, it primarily focused on enhancing model performance and generalization rather than standardizing benchmark implementation procedures. The benchmark study \underline{acknowledged} the complexity of evaluating LLMs in tool-use scenarios but has not made specific attempts to address the inconsistency in benchmark execution methodologies.
	
	\item PromptBench \cite{zhu2024promptbench}: The study did not address the potential variability in implementing its adversarial prompt attacks across different models or laboratories, and it did not attempt to standardize or ensure uniform implementation methodologies for these prompts.
	
	\item AgentBench \cite{liu2023agentbench}: The study introduced AgentBench, a new benchmark designed to evaluate LLMs across a variety of environments, and extensively evaluated 27 LLMs, including both API-based commercial models and open-sourced LLMs. However, the study did not directly address the issue of inconsistency in benchmark implementation across different laboratories. Instead, it focused on establishing a new benchmark and evaluating LLMs within that framework without mentioning efforts to standardize implementation protocols to ensure consistent execution across different research teams.

	\item HaluEval \cite{li2023halueval}: The study utilized ChatGPT to automatically generate hallucinated samples, which is subject to the model's capacity to follow complex instructions for hallucination sampling. Their approach introduced inconsistency in benchmark implementation due to the reliance on a single LLM's capacity, potentially leading to variability in the quality of generated samples. The study \underline{acknowledged} this limitation, noting the quality of hallucinated samples is bounded by ChatGPT's understanding of the instructions, but did not propose specific measures to address this inconsistency directly.
\end{itemize}

\subsection{Slow Test Iteration Time}
\label{subsec:Appendix2-slow-iteration-time}

\textbf{Prevalence}: 18/23

\begin{itemize}
	\item MMLU \cite{hendrycks2020measuring}: The study faced slow test iteration time, evident from its reliance on comprehensive evaluation across a diverse set of subjects and the significant processing periods required, without specific mention of efforts to address this benchmark inadequacy.

	\item FLUE \cite{shah2022flue}: The study introduced benchmarks for evaluating Large Language Models in the financial domain, necessitating extensive testing and iterations across various scenarios, likely extending the time frame for benchmark completion. The study did not explicitly acknowledge this inadequacy or mention attempts to address it.
	
	\item MultiMedQA \cite{singhal2023large}: The study utilized a benchmark combining multiple datasets and evaluated models with human and automated methods, which inherently extends the evaluation timeframe, but did not explicitly mention efforts to address or mitigate slow iteration time.

	\item T-Bench \cite{xu2023tool}: The study demonstrated that the process of enhancing open-source Large Language Models (LLMs) with techniques like model alignment, system prompts, and in-context demonstration retrievers required a practical level of human supervision, indicating that the evaluation of LLMs using the ToolBench benchmark could not be fully automated and might be subject to prolonged iteration time. The study \underline{acknowledged} the challenge of slow iteration time implicitly by emphasizing the practical amount of human supervision needed for enhancing LLMs, although it did not explicitly address measures to reduce evaluation time.
	
	\item Chain-of-Thought Hub \cite{fu2023chain}: The study introduced the Chain-of-Thought Hub, which aimed to evaluate LLMs' reasoning capabilities, but did not specifically address the slow test iteration time. This benchmark required evaluating new models across multiple complex reasoning tasks, which likely extended the evaluation period significantly, similar to the slow iteration time observed in benchmarks like BIG-bench and HELM. However, the study did not acknowledge this inadequacy or mention attempts to address it.
	
	\item KoLA \cite{yu2023kola}: The study introduced a benchmark that requires manual annotation for evolving tasks, which was time-intensive and cannot be automated, leading to potential delays in evaluating LLMs akin to those seen in frameworks like BIG-bench and HELM. The benchmark did not address the need for quicker iteration time, despite mentioning the evolving nature of data and tasks.
	
	\item SciBench \cite{wang2023scibench}: The study introduced SciBench, a benchmark requiring detailed solutions and complex reasoning, involving manual LaTeX formatting from PDFs to ensure data integrity and minimize training data leakage. Despite efforts to mitigate slow iteration time through detailed problem and solution documentation, the study's extensive and manual data processing indicated potential slow test iteration time for evaluating LLMs, acknowledging but not explicitly addressing this benchmark inadequacy.
	
	\item ARB \cite{sawada2023arb}: The study introduced a rubric-based evaluation method for advanced reasoning tasks and tested LLMs, including GPT-4, on a diverse set of problems, requiring human evaluators for complex symbolic answers and proof-like questions, indicating prolonged evaluation time. While it \underline{acknowledged} the challenges in automating the evaluation of advanced reasoning tasks, it did not specifically address measures to significantly reduce test iteration time.

	\item BIG-Bench \cite{srivastava2022beyond}: The benchmark study emphasized the computational expense of full evaluation, especially with programmatic tasks, indicating a lengthy process without addressing the efficiency of iteration time directly.
	
	\item AGIEval \cite{zhong2024agieval}: The study utilized a human-centric benchmark, AGIEval, specifically designed for evaluating LLMs, focusing on tasks derived from official public and high-standard exams. Despite its innovative approach, the study did not explicitly address the potential for slow test iteration time due to the comprehensive and manual nature of its evaluation process, involving extensive human comparison and analysis. This indicates that the process of evaluating new models with AGIEval could be time-consuming, aligning with the described benchmark inadequacy of slow iteration time. There was no mention of attempts to expedite the evaluation process or automate the benchmarking to mitigate this issue.
	
	\item ToolAlpaca \cite{tang2023toolalpaca}: The study utilized a multi-agent simulation to generate a corpus for evaluating LLMs, a process that cannot be automated and likely extends over weeks or months due to manual intervention and complexity, directly mirroring the inadequacy of slow test iteration time. The study did not mention addressing this particular benchmark inadequacy.
	
	\item HELM \cite{liang2022holistic}: The study implemented a comprehensive and systematic evaluation approach to benchmarking LLMs, incorporating extensive metrics and scenarios, which inherently necessitated significant computational resources and time for thorough evaluation, thus likely extending iteration time for model testing. The study, while \underline{acknowledging} the broad and intricate scope of its benchmark, did not specifically address or propose solutions to reduce the slow test iteration time inherent to its extensive evaluation methodology.
	
	\item PromptBench \cite{zhu2024promptbench}: The study required iterating over the entire dataset 100 times on average to generate one adversarial prompt, indicating a potentially prolonged evaluation process, which aligned with the mentioned inadequacy of slow iteration time in benchmarks. The study did not specifically address or propose solutions to reduce these iteration time, suggesting that the challenge of aligning rapid AI development with thorough and effective benchmarking remains unaddressed.
	
	\item AgentBench \cite{liu2023agentbench}: The study developed a comprehensive and systematic benchmark, AgentBench, to evaluate LLMs as agents across a wide array of real-world challenges, requiring multi-turn interactions with detailed and complex environments. This inherently suggested a potentially time-consuming evaluation process, given the variety and depth of the tasks involved. Although the study provided an integrated toolkit to streamline LLM evaluation, the complexity and breadth of the benchmark likely extended the time frame for completion, aligning with the inadequacy mentioned. The study did not specifically address the speed of iteration time as a limitation or concern in its evaluation framework.
	
	\item APIBank \cite{li2023api}: The study introduced API-Bank, a benchmark that requires manual annotation and extensive API integration, which inherently extended the evaluation time frame, echoing the inadequacy of slow test iteration time. Although they employed a novel method to reduce annotation costs, the benchmark's design necessitated detailed and manual interventions, which likely prolong the overall process. The study did not mention any specific strategies to address or mitigate the slow iteration time directly.
	
	\item C-Eval \cite{huang2023c}: The study outlined the creation and evaluation of the C-EVAL suite, emphasizing rapid understanding and improvement of LLM capabilities, but did not specifically address or propose solutions for slow test iteration time inherent in comprehensive evaluations like those mentioned for BIG-bench and HELM. This indicated that the process of using C-EVAL for LLM evaluation could indeed be lengthy and possibly lacked automation, aligning with the mentioned inadequacy. The document did not explicitly acknowledge or attempt to address the slow iteration time challenge.
	
	\item BOLAA \cite{liu2023bolaa}: The study's benchmark, BOLAA, was designed to orchestrate multiple LLM-augmented autonomous agents, and required extensive testing across various scenarios, inherently extending the timeframe for completion. The benchmark did not explicitly address or attempt to reduce the slow iteration time associated with evaluating new models, making it susceptible to the described inadequacy.
	
	\item HaluEval \cite{li2023halueval}: The study utilized a two-step process, sampling-then-filtering, for generating hallucinated samples, which inherently involved a time-consuming, manual annotation component by human labelers. The study \underline{acknowledged} the limitation of slow iteration time, due to the reliance on manual human annotation and complex instruction following by LLMs for quality control, but it did not present a solution to expedite the evaluation process significantly.
\end{itemize}

\subsection{Challenge of Proper Prompt Engineering}
\label{subsec:Appendix2-prompt-engineering-challenge}
\textbf{Prevalence}: 14/23

\begin{itemize}
	\item MMLU \cite{hendrycks2020measuring}: The study designed the benchmark to assess models in zero-shot and few-shot settings across a wide range of subjects, focusing on evaluating knowledge acquired during pretraining without directly addressing the intricacies of prompt engineering. It mentioned the use of various prompts, but did not scrutinize the optimization of such prompts to avoid biases or misinterpretations that could skew assessment results. The study \underline{acknowledged} the limitation of current benchmarks in accurately reflecting models' capabilities, but did not specify efforts to address prompt engineering challenges specifically.
	
	\item LegalBench \cite{guha2024legalbench}: The study outlined challenges in prompt engineering, notably in crafting prompts that accurately assess LLMs' legal reasoning capabilities without introducing bias, thus impacting benchmark integrity. The study \underline{acknowledged} the complexity of legal language and the ongoing effort to refine evaluation techniques, indicating an attempt to address benchmark inadequacies but also highlighting the unresolved nature of prompt engineering challenges.
	
	\item FLUE \cite{shah2022flue}: The benchmark development focused on creating assessments across various NLP tasks in the financial domain, but did not explicitly mention efforts to mitigate or acknowledge the intricacies involved in crafting unbiased and effective prompts to accurately gauge LLM capabilities.
	
	\item MultiMedQA \cite{singhal2023large}: The study \underline{acknowledged} the difficulty in creating prompts that accurately assess LLMs without introducing biases or misinterpretations, affecting both functionality and cybersecurity domains. It attempted to address this through instruction prompt tuning, aiming to align LLMs more closely with medical domain requirements, but the challenge remains complex and unresolved, indicating ongoing issues with prompt engineering adequacy in benchmark assessments.
	
	\item M3KE \cite{liu2023m3ke}: The study utilized a unified prompt for all models across different settings without detailing efforts to mitigate biases or inaccuracies in prompt formulation, impacting the models' evaluation accuracy and potentially overlooking the complexity of assessing LLMs' true capabilities. The study did not explicitly acknowledge or address this benchmark inadequacy.
	
	\item T-Bench \cite{xu2023tool}: The study developed ToolBench, a benchmark to evaluate open-source LLMs for tool manipulation tasks, employing techniques like model alignment, demonstration retrieval, and generation regulation with system prompts. However, it did not explicitly address the intricacies of crafting unbiased and effective prompts to accurately measure a model's capabilities. The study \underline{acknowledged} the importance of prompt engineering by incorporating system prompts designed to guide model generation, yet it did not detail efforts to address or mitigate the potential biases and limitations inherent in prompt design.
	
	\item Chain-of-Thought Hub \cite{fu2023chain}: The study introduced the Chain-of-Thought Hub to measure reasoning capabilities of LLMs using a suite of reasoning benchmarks without addressing the intricacy of crafting prompts that accurately assess these capabilities without bias, which was crucial for fair evaluation. The study \underline{acknowledged} the challenge of evaluating complex reasoning capabilities in LLMs, but did not specifically mention attempts to address the prompt engineering 
	
	\item ARB \cite{sawada2023arb}: The study introduced a novel benchmark that included advanced reasoning problems and proposed a rubric-based self-evaluation method for assessing LLMs, \underline{acknowledging} the difficulty in prompt engineering, but did not provide a conclusive solution to this challenge.
	
	\item BIG-Bench \cite{srivastava2022beyond}: The study \underline{acknowledged} the difficulty in crafting prompts that accurately reflected a model's capabilities without introducing biases or misinterpretations. However, it did not specifically address or propose solutions to overcome those challenges in prompt engineering, highlighting an ongoing inadequacy in effectively evaluating LLMs through benchmarks.
	
	\item HELM \cite{liang2022holistic}: The study utilized a standardized few-shot prompting adaptation for all models, indicating an awareness of prompt engineering challenges. However, it also highlighted the sensitivity of model performance to prompt formatting and adaptation methods, revealing an ongoing struggle with crafting prompts that accurately assess model capabilities without introducing biases or inaccuracies. The benchmark acknowledged this inadequacy by discussing the variation in model performance based on different prompting strategies.
	
	\item AgentBench \cite{liu2023agentbench}: The study designed and implemented AgentBench to evaluate LLMs as agents across various environments, including code-grounded, game-grounded, and web-grounded scenarios, without explicitly addressing the challenges of crafting unbiased and representative prompts that accurately assess a model's capabilities. This omission suggested that the study might not fully account for the complexities of prompt engineering, potentially affecting the accuracy of its evaluations. The benchmark study \underline{acknowledged} the need for systematic evaluation of LLMs as agents, but did not specifically address or attempt to mitigate the inadequacies associated with prompt engineering.
	
	\item C-Eval \cite{huang2023c}: The study implemented a comprehensive evaluation of LLMs on C-EVAL, including both answer-only and chain-of-thought settings, without explicitly addressing or mitigating the intricacies of prompt engineering. Although the study detailed the creation and application of C-EVAL for evaluating LLMs across various disciplines and difficulty levels, it did not discuss specific measures to ensure the prompts accurately and effectively elicit the models' capabilities without introducing biases or misinterpretations. The benchmark study recognized the importance of evaluating advanced abilities of LLMs in a Simplified Chinese context but did not explicitly mention efforts to address the prompt engineering challenge.
	
	\item BOLAA \cite{liu2023bolaa}: The study did not specifically address or attempt to mitigate the intricacies of prompt engineering for accurately and effectively eliciting the capabilities of language models. Instead, it focused on evaluating the performance of various LAA architectures and their orchestration without delving into the effects of prompt design on benchmark outcomes. The study \underline{acknowledged} the importance of prompt engineering indirectly by experimenting with different LAA architectures and their interaction strategies, but did not offer a solution to the inadequacy of prompt engineering itself.
	
	\item HaluEval \cite{li2023halueval}: The study \underline{acknowledged} the challenge of prompt engineering by designing a two-step generation and evaluation process to generate and evaluate hallucinated responses, but it did not explicitly address the inadequacy of ensuring prompts accurately reflect LLM capabilities without introducing biases.
\end{itemize}

\newpage

\section{Examples of Benchmark Inadequacies in Human Dynamics}
\label{sec:Appendix3}
\setcounter{subsection}{0}

\subsection{Diversity in Human Curators and Evaluators}
\label{subsec:Appendix3-diversity-human-curators-evaluators}
\textbf{Prevalence}: 19/23

\begin{itemize}
	\item MMLU \cite{hendrycks2020measuring}: The study designed a benchmark that relied on the aggregation of questions from various sources without explicitly addressing the potential biases introduced by the diversity of human curators and evaluators. It \underline{acknowledged} the importance of diverse evaluative perspectives, but did not detail measures to address this benchmark inadequacy directly.
	
	\item HumanEval \cite{chen2021evaluating}: The study utilized an evaluation set called HumanEval, composed of hand-written programming problems assessed through automated unit tests, without explicit mention of human evaluators in the benchmarking process. The study did not address the potential variability and subjectivity that human evaluators could introduce, focusing instead on automated correctness checks.
	
	\item LegalBench \cite{guha2024legalbench}: The study developed LegalBench with tasks contributed by legal professionals, which did not explicitly address the diversity in backgrounds of these contributors or evaluators, possibly leading to the benchmark not uniformly assessing LLMs. The benchmark study \underline{acknowledged} the collaborative nature of task contributions but did not specifically address the diversity aspect of human curators and evaluators.

	\item MultiMedQA \cite{singhal2023large}: The study engaged clinicians and laypeople from diverse locations (USA, UK, India) for the evaluation of LLM outputs, highlighting variability in human judgments due to their backgrounds. However, it \underline{acknowledged} this diversity and attempted to address it by using a panel of clinicians and lay users for evaluations, aiming for a comprehensive understanding across different demographics.
	
	\item M3KE \cite{liu2023m3ke}: The study involved human-generated multiple-choice questions covering a broad range of subjects and educational levels without explicitly addressing the variability in cultural, religious, political, and academic backgrounds. The benchmark study did not acknowledge or attempt to address this aspect of benchmark inadequacy.
	
	\item T-Bench \cite{xu2023tool}: The study involved human curators in the creation of demonstration examples and the alignment of models with programmatic data, demonstrating a reliance on human judgment for benchmark development. However, the study did not explicitly address the potential biases or inconsistencies introduced by this human involvement, nor did it outline specific measures taken to mitigate such issues.
	
	\item Chain-of-Thought Hub \cite{fu2023chain}: The study curated benchmarks without addressing the diversity in human curators and evaluators, leading to potential biases in benchmark development and evaluation. The benchmarks did not account for the variability in cultural, religious, political, and academic backgrounds of humans, which could introduce inconsistencies and subjectivity in LLM evaluation. The study did not acknowledge or attempt to address this inadequacy.
	
	\item KoLA \cite{yu2023kola}: The study introduced the KoLA benchmark, focusing on evaluating LLMs' world knowledge across various cognitive abilities without specifically addressing the diversity of human curators and evaluators. While it emphasized unbiased and fair evaluation through known and evolving data sources, it did not explicitly mention efforts to mitigate or acknowledge the influence of human diversity in its creation or evaluation processes, potentially leaving room for subjectivity and inconsistency in benchmark development and interpretation.
	
	\item SciBench \cite{wang2023scibench}: The study involved human annotators in data collection and verification, emphasizing the influence of human diversity on the benchmark's development and evaluation. It \underline{acknowledged} the challenges of diverse human interpretation, but did not detail efforts to address this specific benchmark inadequacy.
	
	\item Xiezhi \cite{gu2023xiezhi}: The study employed manual annotations and used ChatGPT for tagging, introducing subjectivity and potential inconsistency due to the reliance on human interpretation and machine understanding, without addressing the diversity of evaluators' backgrounds. The study \underline{acknowledged} the importance of diverse domain knowledge, but did not explicitly address the potential biases introduced by the human element in the creation and evaluation of the benchmark.
	
	\item BIG-Bench \cite{srivastava2022beyond}: The study engaged a team of expert raters to complete tasks submitted to BIG-bench, employing those human evaluators without addressing the potential variability and bias introduced by their diverse cultural, religious, political, and academic or commercial backgrounds. While aiming for a strong human rater baseline, the study \underline{acknowledged} the challenge of representing \enquote{human performance} due to the wide-ranging content within BIG-bench, but did not provide specific strategies to address the diversity-related inadequacies in benchmark evaluation.
	
	\item AGIEval \cite{zhong2024agieval}: The study \underline{acknowledged} the challenge of subjectivity in human evaluation, but did not provide a specific solution to address the inconsistency in benchmarks due to diverse human interpretations, especially in evaluating LLMs using a human-centric benchmark derived from various high-standard exams.
	
	\item ToolAlpaca \cite{tang2023toolalpaca}: The study utilized ChatGPT as a user agent to generate instructions and GPT-3.5 as an assistant agent for structured output generation, relying on human-curated inputs and annotations for evaluating model performance, which could be influenced by the heterogeneity of human backgrounds and subjective interpretations. The study \underline{acknowledged} the importance of diversity in its dataset construction and aimed to mitigate this by generating a diverse corpus. However, it did not specifically address the potential biases introduced by human evaluators' diversity in the development and evaluation of benchmarks.
	
	\item ToolBench \cite{qin2023toolllm}: The study used ChatGPT for the construction of the ToolBench benchmark, involving the generation of diverse instructions and solution paths for real-world APIs, inherently relying on the human-like reasoning capabilities of ChatGPT for curating and evaluating benchmarks. The study did not explicitly address the potential biases or inconsistencies that might arise from the diverse backgrounds of human curators (in this context, the ChatGPT model's training data), nor did it attempt to mitigate the influence of such diversity on the benchmark's development and evaluation.
	
	\item PromptBench \cite{zhu2024promptbench}: The study did not address the diversity of human curators and evaluators, focusing instead on generating adversarial prompts across multiple levels to test LLMs' robustness, which did not directly relate to or account for the variability in human interpretation and judgment based on cultural, religious, political, and academic or commercial backgrounds.
	
	\item AgentBench \cite{liu2023agentbench}: The study implemented AgentBench, a benchmark that evaluated LLMs across various environments without explicitly addressing or attempting to mitigate the diversity in human curators and evaluators, which could lead to inconsistencies in benchmark development and evaluation due to the subjectivity introduced by human involvement.
	
	\item APIBank \cite{li2023api}: The study utilized human annotators for dialogue annotation, which inherently introduced the variability in cultural, religious, political, and academic or commercial backgrounds of these humans. This process led to inconsistencies in the benchmarks' development and their evaluation. Although the study aimed to ensure the quality of annotations through discussion and review by multiple annotators, it did not specifically address the potential bias and inconsistency arising from the diverse backgrounds of these human evaluators.
	
	\item C-Eval \cite{huang2023c}: The study involved human validation in the development of C-EVAL, indicating subjectivity and potential inconsistency due to the diverse backgrounds of the validators, yet did not address this aspect of benchmark inadequacy directly.
	
	\item HaluEval \cite{li2023halueval}: The study involved human labelers in annotating hallucinations in LLM responses, indicating a reliance on diverse human evaluators which can introduce variability in benchmark outcomes due to their different backgrounds. The study \underline{acknowledged} the role of human annotators, but did not address the potential inconsistency their diversity might introduce in evaluating LLMs.
\end{itemize}

\subsection{Diverse Cultural, Social, Political, Religious and Ideological Norms}
\label{subsec:Appendix3-cultural-ideological-challenges}

\textbf{Prevalence}: 18/23

\begin{itemize}
	\item MMLU \cite{hendrycks2020measuring}: The benchmark, by design, evaluated LLMs across a broad spectrum of subjects including ethics, law, and societal norms, without specifically addressing or compensating for the diversity of cultural and ideological perspectives. Their approach might not fully capture the pluralistic nature of human beliefs and values, especially in tasks involving ethical decision-making and societal norms, indicating a potential inadequacy in respecting diverse perspectives through standardized answers or rubrics. The study \underline{acknowledged} the complexity of evaluating LLMs on socially relevant subjects like morality and law, but did not explicitly address attempts to overcome this inadequacy.
	
	\item LegalBench \cite{guha2024legalbench}: The study developed benchmarks for legal reasoning without explicitly addressing or adapting to diverse cultural and legal norms across different jurisdictions, which could lead to standardized answers that do not accommodate the variety of legal, cultural, and ideological perspectives worldwide. The study did not acknowledge or attempt to address this form of benchmark inadequacy.
	
	\item FLUE \cite{shah2022flue}: The study introduced benchmarks in the financial domain without detailed consideration of cultural and ideological diversity in its standardized evaluation metrics, which could conflict with values like diversity and inclusivity, particularly in tasks such as sentiment analysis and news classification that inherently required cultural sensitivity. The study did not explicitly acknowledge or address this benchmark inadequacy.
	
	\item MultiMedQA \cite{singhal2023large}: The study \underline{acknowledged} the complexity of accurately representing diverse cultural and ideological norms in benchmarking LLMs for clinical applications, but did not provide a resolved methodology for integrating these aspects into the benchmarks. The study's focus on creating a diverse benchmark, MultiMedQA, aimed to evaluate LLMs across various medical question-answering datasets, including those involving consumer medical questions, yet it did not specifically address how to incorporate or evaluate the pluralistic nature of human beliefs and values directly within the benchmark's design or evaluation criteria.
	
	\item M3KE \cite{liu2023m3ke}: The study \underline{acknowledged} the challenge of integrating a broad spectrum of viewpoints by covering a wide variety of subjects including humanities, politics, law, and religion, aiming for inclusivity and cultural sensitivity in its benchmarks. However, it did not explicitly address the method of reconciling divergent beliefs and values, nor did it mention any attempts to adjust the standardized answers or rubrics for cultural and ideological diversity, leaving this inadequacy unresolved.
	
	\item Chain-of-Thought Hub \cite{fu2023chain}: The study focused on evaluating the reasoning capabilities of LLMs across various benchmarks, with no specific mention or emphasis on addressing the integration of diverse cultural and ideological norms in the benchmarks used. The benchmarks aimed to assess LLMs' reasoning abilities without adequately considering the diversity of cultural, religious, and ideological perspectives that could influence the interpretation of questions or the appropriateness of answers. This oversight implied that the benchmarks might not fully capture or respect the pluralistic nature of human values and beliefs, potentially leading to biased or non-inclusive evaluations of LLM capabilities. The study did not explicitly acknowledge this limitation nor did it describe efforts to address the integration of diverse perspectives into the benchmarking process. 
	
	\item KoLA \cite{yu2023kola}: The KoLA benchmark, designed to evaluate the knowledge-oriented capabilities of LLMs, incorporates both known and evolving data sources, including Wikipedia and newly published articles, to assess LLMs' understanding and application of knowledge. However, it did not explicitly address the integration of diverse cultural and ideological norms in its evaluation criteria, which could lead to standardized answers or rubrics that may not fully represent the pluralistic nature of human beliefs and values. The study \underline{acknowledged} the challenges in ensuring fairness and reducing biases in LLM evaluations, but did not specify attempts to address the complexities of diverse cultural and ideological norms directly.
	
	\item SciBench \cite{wang2023scibench}: The study demonstrated a reliance on standardized approaches for benchmarking LLMs through complex scientific problem-solving tasks without explicitly addressing or incorporating diverse cultural and ideological considerations into the evaluation process or the construction of benchmarks. The benchmarks were designed without acknowledging the pluralistic nature of human beliefs and values, especially in scenarios requiring ethical decision-making or societal appropriateness, which could lead to biases or skewed results that did not reflect the varied values present in different societies. The benchmark study did not acknowledge this inadequacy of LLM benchmarks, nor made attempts to address it.
	
	\item ARB \cite{sawada2023arb}: The ARB benchmark prioritized problems from domains requiring high-level reasoning without specifically addressing the diverse cultural and ideological norms challenge; it focused on evaluating LLMs' expert reasoning in quantitative subjects and law, without mention of integrating diverse perspectives or the potential for standard answers to conflict with varied societal values. The study did not discuss attempts to address this benchmark inadequacy.
	
	\item Xiezhi \cite{gu2023xiezhi}: The study \underline{acknowledged} this benchmark inadequacy by creating subsets of questions less sensitive and less China-centric, aiming for a more balanced and culturally inclusive assessment, yet it did not fully resolve the challenge of integrating a broad spectrum of cultural and ideological norms.
	
	\item BIG-Bench \cite{srivastava2022beyond}: The benchmark study included tasks such as \enquote{social reasoning}, \enquote{emotional understanding}, and \enquote{figurative language}, which inherently required interpretations that could vary significantly across cultures, yet the study did not explicitly address how these diverse interpretations were considered or integrated into the benchmarking process. This omission indicated that the benchmark might not have fully accounted for or respect the broad spectrum of human diversity in cultural and ideological norms.
	
	\item AGIEval \cite{zhong2024agieval}: The study addressed the challenge of diverse cultural and ideological norms by evaluating LLMs using benchmarks derived from high-standard, official human-centric exams across various domains, and in both English and Simplified Chinese, aiming to capture a broad spectrum of human beliefs and values. However, it did not explicitly mention efforts to address or resolve the benchmark inadequacy of failing to integrate a broad spectrum of cultural and ideological viewpoints through standardized answers or rubrics.
	
	\item HELM \cite{liang2022holistic}: The study's extensive evaluation framework incorporated metrics such as fairness, bias, and toxicity across a diverse set of scenarios, including multiple languages and cultural contexts, aiming to measure LLMs' performance holistically. However, the study \underline{acknowledged} its limitations in fully capturing the pluralistic nature of human beliefs and values, due to inherent challenges in standardizing benchmarks that could respect and represent a broad spectrum of cultural and ideological norms.

	\item PromptBench \cite{zhu2024promptbench}: The study focused on creating adversarial prompts to test LLM robustness but did not address the integration of diverse cultural and ideological norms, which could affect standardized answers' validity across different cultures and ideologies. The benchmarks were designed without consideration for diverse cultural and ideological perspectives, which could lead to biases or skewed results not reflective of varied human values.
	
	\item AgentBench \cite{liu2023agentbench}: The AgentBench framework, designed to evaluate LLMs as agents across various environments, did not explicitly address the integration of a broad spectrum of religious, ideological, legal, political, and cultural viewpoints in its benchmarking process. This omission suggested a potential for standardized answers or rubrics that might not fully respect or represent the pluralistic nature of human beliefs and values, especially in tasks requiring ethical decision-making, societal norms, and cultural interpretations. The study did not mention any acknowledgment or attempts to address this form of inadequacy in its benchmark design.
	
	\item APIBank \cite{li2023api}: The study designed a benchmark that relies on standardized answers or rubrics for evaluating LLMs using API calls without explicitly addressing the incorporation of diverse cultural and ideological viewpoints, which could lead to biases or skewed results not reflective of varied human values. The benchmark did not acknowledge or attempt to address this particular inadequacy.
	
	\item C-Eval \cite{huang2023c}: The study explicitly addressed the challenge of incorporating a wide range of cultural and ideological perspectives within its benchmarks, recognizing the diversity inherent in Chinese society and the subjects covered, which included humanities and social sciences among others. The benchmark's design reflected an understanding that standardized answers might not capture the full spectrum of human diversity, evident in its effort to include diverse disciplines and difficulty levels tailored to the Simplified Chinese context, thus acknowledging the complexity of cultural and ideological norms, without directly stating attempts to resolve this inadequacy.

	\item HaluEval \cite{li2023halueval}: The study implemented a benchmark that did not adequately consider the diverse cultural and ideological norms in its evaluation of LLM hallucinations, which could inherently reflect biases or cultural insensitivities due to the standardized nature of its evaluation criteria. The study did not indicate any acknowledgment of this benchmark inadequacy nor attempts to address it.
\end{itemize}

\end{document}